\documentclass[twocolumn]{svjour3}          
\smartqed  
\usepackage{algorithm, algorithmic}
\usepackage{enumerate}
\usepackage{amsmath,amssymb}
\usepackage{appendix}
\usepackage{graphicx}
\usepackage{wrapfig}
\usepackage{multirow}
\usepackage{cite}
\begin{document}

\title{A Cooperative Group Optimization System
}


\author{Xiao-Feng Xie  \and
        Jiming Liu  \and
        Zun-Jing Wang
}

\authorrunning{Xie, Liu, and Wang} 

\institute{Xiao-Feng Xie \at
              The Robotics Institute, Carnegie Mellon University, Pittsburgh, PA 15213\\
              Tel.: +1-412-2687358\\
              \email{xfxie@alumni.cmu.edu, xie@wiomax.com}           
           \and
          Jiming Liu \at
              Department of Computer Science, Hong Kong Baptist University, Hong Kong \\
              \email{jiming@comp.hkbu.edu.hk}           
           \and
          Zun-Jing Wang \at
              Department of Physics, Carnegie Mellon University, Pittsburgh, PA 15213 \\
              \email{zwang@cmu.edu, wang@wiomax.com}           
}

\date{}

\maketitle

\begin{abstract}
\label{pap:abstract}
A cooperative group optimization (CGO) system is presented to implement CGO cases by integrating the
advantages of the cooperative group and low-level algorithm portfolio design.
Following the nature-inspired paradigm of a cooperative group, the agents not only explore in a parallel way with their individual memory, but also cooperate with their peers through the group memory. Each agent holds a portfolio of (heterogeneous) embedded search heuristics (ESHs), in which each ESH can drive the group into a stand-alone CGO case, and hybrid CGO cases in an algorithmic space can be defined by low-level cooperative search among a portfolio of ESHs through customized memory sharing. The optimization process might also be facilitated by a passive group leader through encoding knowledge in the search landscape. Based on a concrete framework, CGO cases are defined by a script assembling over instances of algorithmic components in a toolbox. A multilayer design of the script, with the support of the inherent updatable graph in the memory protocol, enables a simple way to address the challenge of accumulating heterogeneous ESHs and defining customized portfolios without any additional code. The CGO system is implemented for solving the constrained optimization problem with some generic components and only a few domain-specific components. Guided by the insights from algorithm portfolio design, customized CGO cases based on basic search operators can achieve competitive performance over existing algorithms as compared on a set of commonly-used benchmark instances. This work might provide a basic step toward a user-oriented development framework, since the algorithmic space might be easily evolved by accumulating competent ESHs.

\end{abstract}

\section{Introduction}

Under a suitable formulation, an optimization problem can be cast to a search in a \emph{landscape} \cite{Stadler:1999p1049} over a space of states, which is conceptually simple, but often computationally difficult. The paradigm is general from computational, evolutionary and cultural perspectives. 

Over the past few decades, many general-purpose optimization algorithms have been proposed. Single-start examples include hill climbing, simulated annealing, tabu search, and plenty of other stochastic local search heuristics \cite{Hoos2004}. Population-based examples include genetic algorithm (GA) \cite{Deb:2000p1200,Farmani:2003p1127}, evolution strategy (ES) \cite{Runarsson:2005p1196,Wang:2008p1549,MezuraMontes:2005p1325}, memetic algorithm (MA) \cite{Ong2006,Chen2012}, cultural algorithm (CA) \cite{Reynolds:2008p1078,Becerra:2006p1496}, ant colony optimization (ACO) \cite{Socha2008}, particle swarm optimization (PSO) \cite{Kennedy2001,Lu:2008p1507}, differential evolution (DE) \cite{Price:2005p2148,Becerra:2006p1496,Omran:2009p1303}, social cognitive optimization (SCO) \cite{Xie:2002p1415}, group search optimizer (GSO) \cite{He:2009p1043}, and some other algorithms \cite{Takahama:2005p1387,Liu:2007p1475,Xie:2004p1408,Xie:2009p977,Ullah:2009p1063}.

On the one hand, existing algorithms have explored various metaphors, in which evolution and learning \cite{Hinton:1987p217} are central issues for the adaptability in different landscapes. Algorithms inspired by biological evolution, such as GA and ES, indicate the power of emergent collective intelligence at the population level. Both CA and MA try to emulate cultural evolution upon a canonical population: CA \cite{Reynolds:2008p1078} is a dual inheritance system, which uses a belief space to provide positive clues for the population; and MA \cite{Ong2006,Chen2012} stresses that individual learning, which is normally realized by local search heuristics, can guide the evolution \cite{Hinton:1987p217}. From the viewpoint of learning, evolution can be seen as evolutionary learning \cite{Curran:2006p1143}, in which public information can be regarded as a collective memory used by cooperative search entities. In ACO, heuristics are owned by reflex agents called \textit{ants} \cite{Socha2008} without individual memory, which are cooperated on inadvertent public memory.

Groups are very common in animals \cite{Galef:1995p1128,Leonard2012} and human communities \cite{Tomasello:1993p1330,Nemeth:1986p980,Dennis:1993p1298,Goncalo:2006p1071,Paulus:2000p1114}. Well-studied group phenomena include collective cognition \cite{Leonard2012}, cultural learning \cite{Galef:1995p1128,Tomasello:1993p1330,Curran:2006p1143,Boyd2011}, and group intelligence \cite{Goncalo:2006p1071,Paulus:2000p1114,Satzinger1999,Woolley2010}, of which can promote adaptability and productivity. 
From an algorithmic viewpoint, a group can be represented by multiple agents that search the solutions in a common environment \cite{Platon:2007p1243}, in which the problem landscape can be seen as a common metric space associated with a computational or cognitive representation. 
 In a \emph{cooperative group}, each agent possesses a limited search capability through a mix of both individual and social learning \cite{Galef:1995p1128,Boyd2011,Curran:2006p1143,Tomasello:1993p1330}. Compared to a \emph{stigmergic group}, e.g., ACO \cite{Socha2008}, the agents in a cooperative group can preserve some promising minority patterns \cite{Nemeth:1986p980} with their personal memories \cite{Glenberg:1997p1390,Ericsson:1995p1364} while they search in a parallel way. Compared to a \emph{nominal group} \cite{Dennis:1993p1298} that individuals work separately,
the cooperative agents also interact with their peers through the shared group memory \cite{Danchin:2004p1204,Dennis:1993p1298}.
On the other hand, existing algorithms have provided plenty of search components, and hybrid metaheuristics \cite{Blum2011,Talbi:2002p1393,Parejo2012} has been widely used for optimization. In \cite{Ong2006}, local search strategies were adaptively employed. In \cite{Runarsson:2005p1196}, ES was improved by using a differential variation. DE has been hybridized with different algorithms \cite{Zhang:2003p1404,Omran:2009p1303}. In OEA \cite{Liu:2007p1475}, several evolutionary operators searched together. There are some significant practices in multimethod \cite{Vrugt2009}, multi-operator \cite{Elsayed2011,Elsayed2012,Elsayed2013}, and ensemble algorithms \cite{Mallipeddi2010,Mallipeddi2010a,Mallipeddi2010b}.
These algorithms, either in pure or hybrid forms, have been shown to be \emph{competent} over different sets of problem instances, as measured using some quality metrics \cite{Hoos2004}. 

The motivation behind metaheuristic frameworks \cite{Lau:2007p1228,Parejo2012} might be explained using the No Free Lunch theorems \cite{Wolpert:1997p1149} that any algorithm can only be competent on some problem instances. Conceptual frameworks \cite{Raidl2006,Talbi:2002p1393,Milano:2004p1345,Taillard2001} have been proposed for providing common terminologies and classification mechanisms. Typical software realizations include HeuristicLab \cite{Wagner2009}, ParadisEO \cite{Cahon2004}, and JCLEC \cite{Ventura2008}, etc. Within these frameworks, different algorithm paradigms are coded with some specific interfaces, and are then configured using configuration files to pick instances in a toolbox \cite{Raidl2006,Anderson:2005p1258,Gigerenzer:2001p1178} of reusable components. Some frameworks provide basic relay and teamwork hybrids \cite{Talbi:2002p1393,Parejo2012}, and some frameworks include advanced mechanisms, e.g.,  ``request, sense and response'' \cite{Lau:2007p1228} and ``operator graph'' \cite{Wagner2009}, that facilitates rapid prototyping of hybrid metaheuristics. Each framework can provide an algorithmic space, but walking within the space might be inefficiently, since algorithmic components are effective only as they are embedded in certain environments, and no easy hint is available to understand their behavioral changes. For end users, advanced knowledge is needed to adapt the framework to user-specific problem sets \cite{Parejo2012}. 

Theoretic work in algorithm portfolio design has provided two nontrivial insights \cite{Huberman:1997p1159,Streeter:2008p1072}. First, combining some competent strategies into a portfolio may improve the overall performance by exploiting the negative correlation among their individual performance. Second, the performance can be further strengthened through low-level cooperative search among individual algorithms. Thus, any competent algorithm cases become precious knowledge to be accumulated to adapt to changes over time.  
For end users, it is much easier to understand the offline performance of individual algorithms rather than to understand the complex behavior of algorithmic components. 

It is challenging to support cooperative algorithm portfolios in a development framework, though. Traditionally, heterogeneous algorithms might only loosely cooperate through a communication medium \cite{Talbi:2002p1393}. Implementing low-level hybridization of heterogeneous algorithms would often require expert-level modification of framework code in forming meaningful algorithms.

In this paper, a cooperative group optimization (CGO) system is proposed to utilize the synergy between the cooperative group and algorithm portfolio design. This nature-inspired metaphor allows us not only inheriting the adaptability and productivity of a cooperative group, but also possessing the generality to accumulate various search heuristics for existing metaphors. Furthermore, each agent hold a portfolio of heterogeneous \emph{embedded search heuristics} (ESHs), in which each ESH can drive the whole group into a stand-alone CGO case, and hybrid CGO cases can be defined by cooperative search among a set of competent ESHs that share customized memory elements. In addition, the optimization process might also be accelerated by a passive group leader through adaptively shaping the search landscape, if any global features are available. 

Based on a concrete framework, CGO cases are defined by a script assembling over different instances of algorithmic components in a toolbox. 
A multilayer design of the script, with the support of the inherent updatable graph in the protocol among memory elements, enables a possible way to address the challenge of accumulating heterogeneous ESHs with a few algorithmic components, and building customized portfolios without writing any additional code. For end users, it is possible to easily define competent hybrid metaheuristics for specific problem sets using offline performance information of individual ESHs, based the insights from portfolio algorithm design.

The rest of this paper is organized as follows. In Section \ref{sec:cgos}, a generic CGO system is presented in details. In Section \ref{sec:COP}, the CGO system is implemented for solving the constrained optimization problem \cite{Deb:2000p1200} with a few domain-specific components. In Section \ref{sec:exp}, based on a set of well-known benchmark instances \cite{Liang2006,Runarsson:2005p1196}, the process of algorithm portfolio design is demonstrated, and the experimental results of customized CGO cases are compared to that of existing algorithms in literature. In Section \ref{sec:RelWork}, we discuss related work and possible extensions. This paper is concluded in the last section.

\section{CGO System}
\label{sec:cgos}

\begin{figure*} [t]
\centering \includegraphics[width=6.20in]{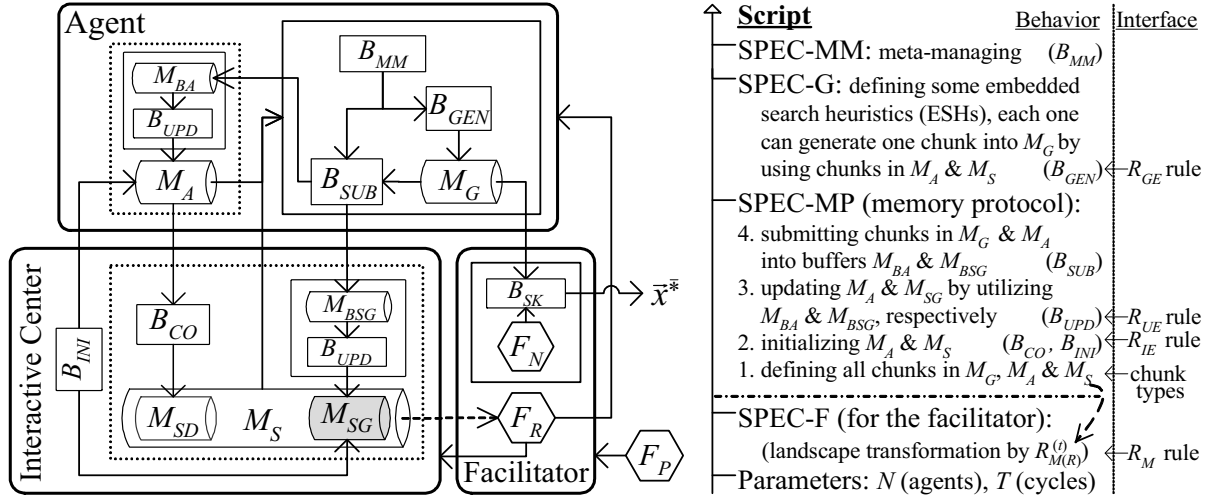} \caption{The CGO system: the multiagent framework, the multilayer script \& primary interfaces, and the toolbox.}
\label{Fig:cgof}
\end{figure*}

The whole CGO system can be represented by a triple, i.e., $<$\emph{Framework}, \emph{Toolbox}, \emph{Script}$>$, as shown in Figure \ref{Fig:cgof}. The toolbox contains some reusable algorithmic \emph{components}. The multiagent framework realizes cooperative group optimization (CGO) algorithms, which is driven by the script with some \emph{interfaces} for embedding valid \emph{instances} of components in the toolbox. Figure \ref{Fig:cgof} is used in the whole section while the details of the CGO system are gradually introduced. 

The system is designed to accommodate three levels of usages. First, the CGO framework supports the basic concept of low-level portfolio algorithm design in a cooperative group. Second, algorithm designers might realize different algorithmic components in the toolbox with some basic interfaces. Finally, basic users can realize (hybrid) CGO algorithms using a multi-layer script, whereas the framework and any components in the toolbox are simply reusable black-box objects. The last two usages also enable the CGO system to be evolvable. 

Basically, the CGO framework follows a modular (and autonomous) design. For a module contains multiple components, an connection to that module means the information can be accessed by all of its components (although might only be used by some of them). For example, $F_{R}^{(t)}$ is accessed by the components in the interactive center and all agents, and $M_A$ and $M_S$ are used by components in the executive module of each agent. There is a direct connection between two components across two modules if the usage is specific. For example, $M_{A}$ of each agent is accessed by $B_{CO}$ in the interactive center. For simplicity, some modules are anonymous.    

This section is structured as follows. We first introduce basic type-based concepts and notations. Sections \ref{sec:CGOF} - \ref{sec:script} then describe the framework, toolbox and script. Further details of memory and behavior in the CGO framework will be described in Section \ref{sec:CGODetails}. In Section \ref{sec:exec}, the execution process of the CGO framework is described based on all these building blocks. 

\subsection{Preliminary Concepts}
\label{sec:kr}

The CGO system is full of knowledge elements that can be organized in a \emph{type}-based representation. 
Each knowledge element can be accessed by using its \emph{identifier}, referring to a \emph{name} and a \emph{type}, in which the type defines some \emph{properties} for facilitating knowledge sharing, and the name ensures the uniqueness. For each type, a \emph{compatible type} is a \emph{subtype} or the same type.  

The general problem-solving capability arises from the interaction of declarative and procedural knowledge \cite{Anderson:2005p1258}. A basic declarative component, called a \emph{chunk}, aggregates a small amount of problem information in a specific data structure. A procedural component contains \emph{actions}, in which each works on some \emph{input}/\emph{output} \emph{parameters}. It is called a \emph{rule} if it has one action. Each component might have some \emph{setting parameters}. For a \emph{macro} component, one or more setting parameters are component types rather than primitive data types.

In the CGO toolbox, each algorithmic component is a binary object that can be instantiated using its \emph{actual type} and valid setting parameters. 
Each parameter or script interface has a \emph{formal type}, which is either a primitive data type or a component type for accepting an instance of any knowledge component if its actual type is a compatible type. 

The CGO script is used for calling instances of algorithmic components of specific \emph{interfaces} in the toolbox. \emph{Primary interfaces} are directly supported types, whereas association interfaces might be introduced from components that are embedded as setting parameters of macro components. 

\subsubsection{Notation}

Normally, a type is notated in the form of $TG_{TK}^{(t)}$ or $TG_{(TT)}^{(t)}$, in which \textit{TG} indicates the \emph{general type}; \textit{TK} represents a \textit{key variant}, e.g., a subtype or with a nontrivial property; \textit{TT} in the subscript parentheses means a simple variant, which is often used as the names of similar instances; and \textit{t} in the superscript parentheses stresses it possesses the dynamic property in a time-varied style, where $t$ means at the $t$th learning cycle. 

Here are general types\footnote{Note that the same symbol no longer means a general type if it appears at other places. Taking ``$M_S$'' as an example (``$M$'' is its general type), ``$S$'' means a key variant rather than the general type of a space of states.} to be used in this paper. Some types are related to the problem, where ``$F$'' means a problem representation and ``$S$'' means a space of states. As major building blocks for solving capability in the CGO framework, ``$M$'' is a \emph{memory} containing some chunks, and ``$B$'' is a \emph{behavior} with actions that directly or indirectly interact with some chunks in memory. ``$C$'' means a setting parameter of a component.

Some general types are used for chunks. ``$CH$'' is used to mention a chunk in general, but each specific chunk has a general type. A set of chunks of the same type ``$CH$'' can be organized into a \emph{chunk set}, called ``$\$CH$'', in which ``$\$$'' means a set. ``$E$'' means a list of ordered chunks of arbitrary types. 

Only the notation for rules is more complicated, since lots of rules, in which actual types can be subtypes of (subtypes of) some formal types, might be realized to make the system flexible and evolvable. 

A rule is notated in the form of $R_{TK(TT)}^{TA(t)}$, in which ``$R$'' is the general type, \textit{TA} stresses an actual type. If \textit{TA} is not used, it is a formal type (an abstract rule) that is used for a parameter. If necessarily, a subtype of $TK$ is notated as \textit{TK}:\textit{TKC}, where \textit{TKC} after ``:'' indicates the unique properties associated with the subtype, and a further subtype can be notated in the same way. 

\subsection{CGO Framework: Overall Description}
\label{sec:CGOF}

The CGO framework supports the cooperative search of a group of totally $N$ \emph{agents}. All the agents are of the same structure. Figure \ref{Fig:cgof} shows one of the agents and the shared environment that contains a \emph{facilitator} and an \emph{interactive center}. Each agent possesses a limited search capability and can only indirectly interact with its peers through the shared environment, in order to achieve the common goal of finding a near optimal solution  $\vec{x}^{\bar{*}}$ for the problem $F_{P}$. 

The CGO framework runs in iterative learning cycles. The execution is terminated if the number of learning cycles ($t$) achieves the maximum cycle number ($T$).

\subsubsection{Facilitator}
\label{sec:facilitator}

For a global optimization problem $F_{P}$ to be solved, an essential landscape \cite{Stadler:1999p1049} can be represented as a tuple $<S_{P}, R_{M}>$. The \emph{problem space} ($S_{P}$) contains all the \emph{states} to be searched, in which each state $\vec{x}$ is a potential solution. For a real-world group, the states of $F_{P}$ might be viewed as creative ideas \cite{Paulus:2000p1114,Satzinger1999}. The \emph{quality-measuring} rule ($R_{M}$) measures the difference of \emph{quality} between $\forall \vec{x}_{(a)} ,\vec{x}_{(b)} \in S_{P} $: if the quality of $\vec{x}_{(a)}$ is better than that of $\vec{x}_{(b)} $, then $R_{M}$($\vec{x}_{(a)} $,$\vec{x}_{(b)}$) returns TRUE, otherwise it returns FALSE. 

The facilitator maintains a \emph{natural representation} ($F_{N}$) and an \emph{internal representation} ($F_{R}$),
which are both formulated from the problem $F_{P}$. 
For $F_{N} = <S_{P}, R_{M:N}>$, the $R_{M:N}$ rule is an $R_{M}$ rule possessing the \emph{natural} property that faithfully measuring the quality among candidate states as the same as in the original problem $F_{P}$. It is used by the \emph{solution-keeping} behavior ($B_{SK}$) to update the \emph{best-so-far} state $\vec{x}^{\bar{*}}$ of $F_{P}$ among all states that are generated by agents. Specifically, $B_{SK}$ replaces $\vec{x}^{\bar{*}}$ by any state $\vec{x} \in S_{P}$ if $\vec{x}$ has a better quality, based on the $R_{M:N}$ rule. The optimal solution of $F_{P}$ is ensured to be kept if it is visited.

For $F_{R} = <S_{P}, R_{M(R)}, AUX>$, the $R_{M(R)}$ rule can be an arbitrary $R_{M}$ rule, and $AUX$ contains any auxiliary components associated with structural information of $F_{P}$ that might be useful for search. The basic usage of $F_{R}$ is to encapsulate any knowledge in $F_{P}$ that will be further processed to find solutions. It is the internal problem used in the interactive center and all agents. 

Both $F_{N}$ and $F_{R}$ are representations on $S_{P}$. However, $F_{N}$ is not used for providing search clues, whereas $F_{R}$ might deviates from original problem landscape during the runtime for facilitate the search process. 

For a specific problem type, $S_{P}$, $R_{M:N}$, and $AUX$ can be predefined, since $S_{P}$ is defined in $F_{P}$, only one $R_{M:N}$ rule is required (since different $R_{M:N}$ rules are equivalent for the usages in $F_{N}$), and $AUX$ is normally rather concise in practical usages (although it is suitable to put any available and useful knowledge into $AUX$). Thus the main effort is to implement an $R_{M}$ rule as the input for $R_{M(R)}$. A simple way is to set $R_{M(R)} = R_{M:N}$, but more domain-specific $R_{M(R)}$ might be designed if landscape features are available. An example of the facilitator will be demonstrated in Section \ref{sec:COP_FAC}.

The facilitator might be viewed as a passive group leader, who can influence the solving process by providing and adaptively updating $F_{R}$, but not directly managing the operations of any agents. The problem landscape might be transformed by using an unnatural $R_{M}$ rule that incorporates suitable knowledge. For example, approximate models \cite{Jin:2002p983} may smooth a rugged landscape, and constraint-handling techniques \cite{Hamida:2002p1205,Xie:2004p1408,Runarsson:2005p1196} have been widely used. An adaptive $R_{M(R)}^{(t)}$ rule might be realized by feeding a run-time chunk in $M_{S}$.

\subsubsection{Agents and Interactive Center}
\label{sec:AgentsIC}

The search process for solving $F_{R}$ is performed by the agents with the support of the interactive center. The general solving capability arises from the interplay between \emph{memory} and \emph{behavior} \cite{Anderson:2005p1258}. 

In cognitive theories, memory \cite{Glenberg:1997p1390,Anderson:2005p1258,Ericsson:1995p1364} is a basic component for supporting the learning process. The importance of using (public) memory has been also addressed in some computational frameworks \cite{Lau:2007p1228,Taillard2001,Milano:2004p1345}.

Specifically, memory is used for storing and retrieving chunks, in which each chunk contains certain particularities of state(s) in $F_{R}$. 
In this paper, a conceptualization model \cite{Glenberg:1997p1390} is used, which only requires a bounded space complexity, as compared to the unbounded memory used in some cognitive architectures \cite{Anderson:2005p1258}. Each memory holds a list of permanent \emph{cells}, in which each cell possesses a unique \emph{cell type} and only stores the chunk of a compatible type as its content.

Two memory types, i.e., \emph{long-term memory} (LTM) \cite{Ericsson:1995p1364} and \emph{buffer}, are classified according to if all stored chunks are cleared at the end of each learning cycle or not. For a LTM, the chunk in each cell must be filled at the initialization stage and is subjected to be updated during the run-time. Note that each LTM cell only keeps the most recently updated chunk.  

From the viewpoint of each agent, there are three basic memories, including a generative buffer ($M_{G}$) and an individual memory ($M_{A}$) of its own and a social memory ($M_{S}$) in the interactive center. For each agent, the two long-term memories, i.e., $M_{S}$ and its $M_{A}$, store all currently available past experience for it, while the buffer $M_{G}$ temporarily stores a chunk that is newly generated by it, in each learning cycle. 

In a LTM, a chunk possesses either the \emph{genuine} or \emph{dependent} property according to if it only contains independent data or it is a specific data structure only designating the references to other chunks. In this paper, we only considered three kinds of LTM: $M_{A}$ only contains genuine chunks, whereas two sub-memories $M_{SG}$ and $M_{SD}$ in $M_{S}$ respectively possess genuine and dependent chunks. Here $M_{SD}$ is used for sharing \emph{non-private} chunks \cite{Liu:2006p1045} in $M_{A}$ of all agents.

All chunks in LTMs must be initialized and might be updated during learning cycles. The memories with genuine chunks, including $M_{A}$ in all agents and $M_{SG}$ in the interactive center, are initialized by using the \emph{initializing behavior} ($B_{INI}$). These genuine chunks are updated in a similar way: chunks in $M_{A}$ and $M_{SG}$ are respectively updated by the $M_{A}$-\emph{updating} ($B_{UA}$) and $M_{SG}$-\emph{updating} ($B_{USG}$) behaviors through respectively using the chunks in the buffers $M_{BA}$ and $M_{BSG}$. The dependent chunks in $M_{SD}$ are initialized by the \emph{collecting behavior} ($B_{CO}$) for collecting non-private chunks in $M_{A}$ of all agents, and are automatically updated if the referring chunks in $M_{A}$ of any agents are changed.

For each agent, its executive module performs the \emph{meta-managing} ($B_{MM}$), \emph{generating} ($B_{GEN}$), and \emph{submitting} ($B_{SUB}$) behaviors at each learning cycle. For a given CGO algorithm case, the $B_{MM}$ behavior probabilistically selects one of its executive rows, in which the generative part is executed by the $B_{GEN}$ behavior for outputting a new chunk into $M_{G}$ by using the inputs in $M_{A}$ and $M_{S}$, whereas the updating list is used by the $B_{SUB}$ behavior for submitting chunks from $M_{A}$ (cloned) and $M_{G}$ into specific buffer cells of $M_{BA}$ and $M_{BSG}$. Each chunk in $M_{G}$ possesses the \emph{solution} property that can export a potential solution $\vec{x}$. The potential solution is exported to the facilitator as a candidate for the best-so-far solution $\vec{x}^{\bar{*}}$.

\subsection{CGO Toolbox}
\label{sec:CGOLib}

The toolbox contains some addable/removable algorithmic components of specific interfaces. Each component can be called symbolically using its identifier and setting parameters (thus the actual realization might be in a black box for end users). Primary interfaces are directly called by the CGO script (in Section \ref{sec:script}), whereas background components of association interfaces might be used in components of primary interfaces, if necessarily. One nontrivial usage of the toolbox is to embed knowledge units that are commonly used in existing optimization algorithms, whereas novel algorithmic components might also be supported, if available.

The chunks in $M_{A}$, $M_{S}$, and $M_{G}$ are of some primary chunk interfaces. In general, a chunk interface is notated as ``$CH$''. Since there are a group of agents, if any type $CH$ is used in $M_{A}$ and $M_{G}$, then $\$CH$ is automatically considered as a primary interface, in which ``$\$$'' means a set of chunks.

Most straightforward primary chunk interfaces include a state $\vec{x}$ and a state set $\$\vec{x}$. For example, $\$\vec{x}$ can be used as a population of individuals in many evolutionary algorithms. In stochastic local search strategies \cite{Hoos2004}, $\vec{x}$ is used as an incumbent solution to be improved. There are some other types in existing algorithms. In ES \cite{Runarsson:2005p1196}, the chunk to be generated is of a combined type $(\vec{x}, \vec{\sigma})$, in which $\vec{\sigma}$ is used for a log-normal distribution. In model-based algorithms, probabilistic models (e.g., a pheromone matrix \cite{Socha2008}) are used, which can be seen as chunks in the public memory.

For the facilitator, only $R_{M}$ is considered as a primary rule interface for realizing $R_{M(R)}$ in $F_{R}$. For constraint optimization, $R_{M}$ can be used for embedding constraint-handling methods \cite{Hamida:2002p1205,Xie:2004p1408,Runarsson:2005p1196}. $R_{M}$ might use one chunk in $M_{S}$ as its input for run-time guidance. 

For the agents and the interactive center, there are three primary rule interfaces, i.e., elemental \emph{initializing} ($R_{IE}$),  \emph{updating} ($R_{UE}$), and \emph{generating} ($R_{GE}$) rules. 

Specifically, $R_{IE}$ instances are used by $B_{INI}$ for initializing $M_{A}$ and $M_{SG}$, $R_{UE}$ instances are used by $B_{UA}$ and $B_{USG}$ for updating $M_{A}$ and $M_{SG}$ by respectively using the buffers $M_{BA}$ and $M_{BSG}$, and $R_{GE}$ instances are used by $B_{GEN}$ for generating new chunks into $M_{G}$.

All the components the facilitator can access $F_{N}$, whereas all the components in the agents and the interactive center can access $F_{R}$ by default. Primary rule interfaces might have various subtypes. Some subtypes are problem-specific, whereas some subtypes are generic across different problem types (e.g., the constrained optimization problem, graph coloring, and traveling salesman problem), by using only generic knowledge in $F_{R}$. Only some generic subtypes are introduced here as examples, whereas some problem-specific subtypes will be introduced in Section \ref{sec:COP}, when we demonstrate the actual implementation on a specific problem type. 

\subsubsection{Background Rules}
\label{sec:bgRules}

In this paper, two \emph{selecting} rules are used in other rules ($R_{UE:X}$ and $R_{GE}$ rules respectively in Sections \ref{sec:euRules} and \ref{sec:COP_RGE}). A \emph{selecting} rule ($R_{SEL}$) chooses one state $\vec{x}_{(O)}$ from a state set $\$\vec{x}_{(I)}$. 

The \emph{greedy} $R_{SEL}$ rule ($R_{SEL}^{G}$) has no setting parameter. It simple returns the best state among the states in $\$\vec{x}_{(I)}$  by comparing to each $\vec{x} \in \$\vec{x}_{(I)}$ one by one, using the $R_{M(R)}$ rule in $F_R$.

The \emph{tournament} $R_{SEL}$ rule ($R_{SEL}^{TS}$) has two setting parameters, i.e., a tournament size $C_{NTS}$, a Boolean quality flag $C_{BQ}$. It is executed as the follows. First, totally $C_{NTS}$ states are selected from $\$\vec{x}_{(I)}$ at random. Second, these states are compared by using the $R_{M(R)}$ rule, and the state with a better quality or a worse quality is kept, if $C_{BQ}$ is TRUE or FALSE, respectively. Finally, the last surviving state is outputted as the selected state $\vec{x}_{(O)}$.

\subsubsection{Elemental Initializing Rule}

The $R_{IE}$ rule has an output chunk $CH_{(I)}$. Each actual $R_{IE}$ rule is used for initializing each chunk in $M_{SG}$ and each chunk set  $\$CH_{(M)}$, in which $CH_{(M)}$ is a chunk in $M_{A}$ of each agent.

The $R_{IE:X}$ subtype is an $R_{IE}$ rule that outputs a state set $\$\vec{x}$ as $CH_{(I)}$. For the \emph{random} $R_{IE:X}$ rule ($R_{IE:X}^{RND}$), each element in $\$\vec{x}$ is randomly generated by $R_{GE}^{RND}$ within the problem space $S_{P}$. 

\subsubsection{Elemental Updating Rule}
\label{sec:euRules}

The $R_{UE}$ rule has two input chunks, i.e., $(CH_{(M)}, CH_{(U)})$, and updates the chunk $CH_{(M)}$. Each $R_{UE}$ rule is used for updating a genuine chunk $CH_{(M)}$ in $M_{A}$ or $M_{SG}$. Here we only consider two basic subtypes.

The $R_{UE:S}$($\vec{x}_{(a)} ,\vec{x}_{(d)}$) rule replaces $\vec{x}_{(a)}$ by $\vec{x}_{(d)}$ in a specific condition. There are two simple $R_{UE:S}$ rules: a) the \emph{direct} $R_{UE:S}$ rule ($R_{UE:S}^{D}$), which replaces unconditionally; and b) the \emph{greedy} $R_{UE:S}$ rule ($R_{UE:S}^{G}$), which carries out the replacement if $R_{M(R)}(\vec{x}_{(a)} ,\vec{x}_{(d)} )\equiv$TRUE. 

The $R_{UE:X}$($\$\vec{x}_{(a)}, \$\vec{x}_{(d)}$) rule forms a new $\$\vec{x}_{(a)}$ by picking some of the states in $\$\vec{x}_{(a)} \cup \$\vec{x}_{(d)}$. Various subtypes of $R_{UE:X}$ are used in evolutionary algorithms for updating the population with new individuals. The \emph{tournament-selection} $R_{UE:X}$ rule ($R_{UE:X}^{TS}$) , which has one setting parameter called $C_{NTW}$, is realized as follows. For each state in $\$\vec{x}_{(b)} $, it replaces one state in $\$\vec{x}_{(a)}$ that is selected by an $R_{SEL}^{TS}$ instance (defined in Section \ref{sec:bgRules}) with $C_{NTS}$=$C_{NTW}$ and $C_{BQ}$=FALSE. 

There are some other subtypes used in existing algorithms. For example, in ACO \cite{Socha2008}, $CH_{(M)}$ is a pheromone matrix and $CH_{(U)}$ is a state set $\$\vec{x}$. 

\subsubsection{Elemental Generating Rule}
\label{sec:RGE}

The $R_{GE}$ rule has an ordered list of input chunks $E_{(IG)}$, and an output chunk $CH_{(OG)}$ that has the solution property. Each actual $R_{GE}$ rule performs the search role for generating a new chunk.

Inputs/outputs of $R_{GE}$ rules might be arbitrarily defined, although most of them are defined in simple forms. Using a list of input chunks in $E_{(IG)}$ allows flexible cooperative search between $R_{GE}$ rules by sharing some chunks. Using a single chunk in $CH_{(OG)}$ enables a simple realization, but without loss of generality.

In many existing algorithms, their search operators can be seen as $R_{GE}$ rules that output $\vec{x}$, although they might use different element(s) in $E_{(IG)}$. An extreme case is for a search rule start from scratch, in which $E_{(IG)}=\varnothing$. For example, a random $R_{GE}$ rule ($R_{GE}^{RND}$) is a generic rule that generates a random state within $S_P$. Some of them only use one element in $E_{(IG)}$. For example, a local search heuristic $R_{LS}$ uses an incumbent state, whereas each ant in ACO \cite{Socha2008} uses a pheromone matrix. Some of them, e.g., PSO, DE, and SCO (as will be introduced in Section \ref{sec:COP_RGE}), use multiple elements in $E_{(IG)}$. Some search operators do use other elements rather than $\vec{x}$ as $CH_{(OG)}$. For example, the chunk to be generated in ES \cite{Runarsson:2005p1196} is of a macro type $(\vec{x}, \vec{\sigma})$ for encoding a log-normal distribution around $\vec{x}$.

Moreover, $R_{GE}$ can be realized in macro forms that support some association interfaces. For example, for an $R_{GE}$ rule that has $E_{(IG)} = \{\$\vec{x}\}$ and $CH_{(OG)} = \vec{x}$, a possible relay form  \cite{Talbi:2002p1393} is a tuple $<R_{SEL}, R_{XS}>$ \cite{Xie:2009p977}, in which $R_{XS}$ is a \emph{recombination} rule that outputs $\vec{x}$ by using two parents $\vec{x}_{(1)}$ and $\vec{x}_{(2)}$ independently selected from $\$\vec{x}$ by the $R_{SEL}$ rule. Furthermore, a mutation rule or a local search rule can be linked for perturbing or improving the output state $\vec{x}$ \cite{Ong2006,Talbi:2002p1393}. 

\subsection{CGO Script}
\label{sec:script}

As shown in Figure \ref{Fig:cgof}, the CGO framework is driven by a script realized in multiple layers. The script body includes overall setting parameters, problem specification (SPEC-F), memory protocol specification (SPEC-MP), generative specification (SPEC-G), and meta-management specification (SPEC-MM). For all the interfaces in the script, instances of components in the toolbox are used. Only a few setting parameters on component instances might be defined as script parameters if they are explicitly assigned. In the practical usage, a CGO algorithm case can be defined by a case identifier ($ID_{C}$) and a few script parameters, based on a given script body that reusing a few existing specifications.

The lowest two layers are quite simple. The overall script parameters include the number of agents ($N$) and the maximum number of learning cycles ($T$).
In the facilitator, the problem specification (SPEC-F) is
\[\text{SPEC-F}=<R_{M(R)}>,\]
since the other elements, i.e., $S_{P}$, $R_{M:N}$, and $AUX$, can be easily predefined for a specific problem type. 

The upper three layers are used for driving all modules in the agents and the interactive center. Each layer contains addable/removable elemental rows for supporting an evolvable property in an algorithmic space.

\subsubsection{Memory Protocol Specification}
\label{sec:MPSpec}
The memory protocol specification (SPEC-MP) defines how will the chunks be initialized and updated in $M_{A}$ of all agents and $M_{S}$ of the interactive center, given any chunk is newly generated in $M_{G}$ of the agents. SPEC-MP might be seen as a domain ontology for encoding low-level knowledge units \cite{Edgington:2004p1067}. In Figure \ref{Fig:cgof}, it is used for driving the modules in all agents and the interactive center, except for the \emph{meta-managing} ($B_{MM}$) and \emph{generating} ($B_{GEN}$) behaviors in the agents. 

Formally, SPEC-MP contains a table of \emph{memory protocol rows}, in which each contains five elements, i.e., 
\[\text{SPEC-MP Row}=<ID_{M},CH_{M},R_{IE},R_{UE},CH_{U}>,\]
where $ID_{M} \in \{M_{A}, M_{SG}, M_{SD}\}$, $CH_{M}$ at each row is a unique chunk in the memory referred by $ID_{M}$. In SPEC-MP, the second column defines the three lists of chunks in $M_{A}$, $M_{SG}$, and $M_{SD}$. The list of chunks in the last column contains all chunks in $M_{A}$ and $M_{G}$. The chunks in the columns of $CH_{M}$ and $CH_{U}$ belong to primary chunk interfaces. Each chunk in $M_{G}$ possesses the \emph{solution} property that can export a state $\vec{x}$.

The last three elements are defined differently for genuine and dependent chunks. If $CH_{M} \notin M_{SD}$, $R_{IE}$ and $R_{UE}$ are elemental \emph{initializing} and \emph{updating} rules for $CH_{M}$, and $CH_{U}\in M_{A} \cup M_{G}$ is a candidate chunk for updating $CH_{M}$. If $CH_{M} \in M_{SD}$, $R_{IE}$ and $R_{UE}$ are \emph{null}, and $CH_{U}\in M_{A}$, since each chunk in $M_{SD}$ is automatically updated using chunks in $M_{A}$ of all agents.

The validity of SPEC-MP can be locally checked in two steps. The first step is to ensure that the types used in each row are locally compatible. Notice the fact that there are multiple agents but only one interactive center. If $CH_{M} \in M_{SD}$, then its type is $\$CH_{U}$, in which each element is $CH_{U}$ in $M_{A}$ of all agents. Table \ref{Tab:iotypes} gives the types of input/output parameters of $R_{IE}$ and $R_{UE}$ for the chunks in $M_{A}$ and $M_{SG}$. There are two special cases of using a chunk set. If $CH_{M} \in M_{A}$, $R_{IE}$ is used for initializing $CH_{M}$ in  $M_{A}$ of all agents. If $CH_{M} \in M_{SG}$, $M_{BSG}$ will collect all chunks that are submitted from agents in each learning cycle as the inputs of $R_{UE}$ for updating $CH_{M}$.

\begin{table} [htbp]
\centering \caption{Generic types of input/output parameters of $R_{IE}$ and $R_{UE}$ rules}
\begin{tabular}{|l|c|c|c|} \hline 
$ID_{M}$ & $CH_{(I)}$ of $R_{IE}$ & $CH_{(M)}$ of $R_{UE}$ & $CH_{(U)}$ of $R_{UE}$ \\ \hline 
$M_{A} $ & $\$CH_{M}$ & $CH_{M}$ & $CH_{U}$ \\ \hline 
$M_{SG} $ & $CH_{M}$ & $CH_{M}$ & $\$CH_{U}$ \\ \hline 
\end{tabular}
\label{Tab:iotypes}
\end{table}

The second step is to ensure the validity across all rows. First, each $CH_{M}$ must be unique, and each $CH_{U} \in M_{G}$ must possess the solution property. Second, each chunk $CH_{M}$ should be \emph{updatable}, i.e., has the probability to be updated by chunks that are generated in $M_{G}$, across multiple cycles. Notice that if a row is used in a cycle, $CH_{M}$ is updated by $CH_{U}$.

The validity of updatable relations can be easily checked by using an \emph{updatable graph} that is formed from all rows in SPEC-MP: For each row, $CH_{U}$ is the parent node of $CH_{M}$.
A valid updatable graph contains separate trees, where for each tree, the root is a chunk in $M_{G}$, the children are chunks in $M_{A}$ and $M_{S}$, and each chunk in $M_{S}$ is always a leaf node. An example of a valid updatable graph, which contains a single updatable tree, is provided later in Table \ref{tab:SPEC-MP}. 

Furthermore, SPEC-MP can be easily maintained by using the updatable graph. Except for the root nodes, each other node in the updatable graph is $CH_{M}$ in a unique memory protocol row. Each leaf node (i.e., the corresponding row), if it is not used by the upper layer, can be removed without changing the validity of the remaining graph. A whole tree is removed from the graph if only its root node is left. Each new node can be added to either a leaf node or a root node.

SPEC-MP provides an essential support for the stability in cooperative search. If $CH_{M}^{(t)}$ is used by different search heuristics, it is always updated by $R_{UE}$ and $CH_{U}^{(t)}$ in the same row. 

\subsubsection{Generative Specification}
\label{sec:SPECG}

The generative specification (SPEC-G) contains a set of \emph{generative rows}, in which each generative row is 
\[
\text{SPEC-G Row} = <ID_{G}, R_{GE}, E_{IG}, CH_{OG}>, 
\]
where $ID_{G}$ is a unique name, $R_{GE}$ is an elemental generating rule, $E_{IG}$ is an ordered list of chunks, $CH_{OG}$ is a chunk. Each $ID_{G}$ designates an ESH that is corresponding to a stand-alone algorithm case.

The validity of SPEC-G only needs to be locally ensured in each ESH, based on the memory protocol specification (SPEC-MP). First, each chunk in $E_{IG}$ belongs to $M_{A} \cup M_{S}$, and $CH_{OG}\in M_{G}$. Second, $E_{IG}$ and $CH_{OG}$ are respectively linked to the input/output parameters of the $R_{GE}$ rule. Third, all chunks in $E_{IG} \cup CH_{OG}$ must be updatable, i.e., these chunks form a \emph{subtree} that contains one root node (i.e., a chunk in $M_{G}$) in the updatable graph defined in Section \ref{sec:MPSpec}. 

Each generative row is independent to the other. New ESHs can be freely added into SPEC-G, or be removed from SPEC-G if it is not used in any portfolio. 

\subsubsection{Meta-Management Specification}

The meta-management specification (SPEC-MM) defines a customized portfolio within an algorithm space that is formed by using some ESHs as the bases. 

Each algorithm case has a name ($ID_{C}$) and contains a set of \emph{executive rows}, in which each executive row contains three elements, i.e., 
\[
\text{SPEC-MM Row} = <ID_{G}, E_{UPD}, C_{W}>,
\]
in which $ID_{G}$ is the name of an ESH in SPEC-G, the updating list $E_{UPD}$ contains a list of genuine chunks, and $C_{W} \geq 0$ is a weight value for the row. 

The selection probability for each executive row is $C_{W}/\sum{C_{W}}$. Thus any executive row is ignored if it has $C_{W}=0$. Any two executive rows are called \emph{cooperative} rows if their $E_{UPD}$ lists share at least one element, otherwise they are \emph{independent} to each others.

For ensuring the validity of each CGO algorithm case, $E_{UPD}$ in each row must satisfy $E_{IGG}\subseteq E_{UPD}\subseteq E_{IGM}$, in which $E_{IGG}$ contains the list of all genuine chunks in $E_{IG}$ of the corresponding ESH, and $E_{IGM}=\bigcup{E_{IGG}}$ for $E_{IGG}$ of all ESHs. The intuition is that each chunk that is used by ESHs must be actively updated. By default, there are $C_{W} = 1$ and $E_{UPD} = E_{IGG}$ for each (independent) ESH that is newly added as an executive row in SPEC-MM. 

If ESHs use input chunks in the same tree of the updatable graph in SPEC-MP, their $E_{UPD}$ lists might be customized in $E_{IGM}$. An algorithm case is called as in the \emph{customized} or \emph{default} mode according to any $E_{UPD}$ is customized or not. The customization of $E_{UPD}$ lists can turn independent ESHs into cooperative ESHs, and can make cooperative ESHs cooperating more.

\subsection{CGO Framework: Memory and Behavior}
\label{sec:CGODetails}

As shown in Figure \ref{Fig:cgof}, the memory and behavior in the agents and the interactive center are driven by the script using some components in the toolbox. 

\subsubsection{Long-Term Memory and Buffer Modules}
\label{sec:mem}

As described in Section \ref{sec:MPSpec}, the genuine chunks in $M_{A}$ and $M_{SG}$, and dependent chunks in $M_{SD}$ are defined in the first two columns of SPEC-MP, and the root nodes in the updatable graph of SPEC-MM form the list of chunks that are supported in $M_{G}$. Each chunk in $M_{A}$ and $M_{S}$ can be retrieved by each agent. Any new solution contained in $M_{G}$ is submitted to the facilitator.

The buffers $M_{BA}$ and $M_{BSG}$ are respectively used for updating $M_{A}$ and $M_{SG}$, where their cells are of one-to-one mapping based on each row of SPEC-MP, i.e., each cell in a buffer accepts $CH_{U}$ if it is used for updating a chunk $CH_{M}$ in $M_{A}$ or $M_{SG}$. As shown in the last column of Table \ref{Tab:iotypes}, the corresponding \emph {cell types} of $M_{BA}$ and $M_{BSG}$ are respectively $CH_{U}$ and $\$CH_{U}$, since each agent only submit once to its $M_{BA}$, whereas all agents might submit chunks to cells in $M_{BSG}$.

\subsubsection{Initializing and Updating Behavior}
\label{sec:bini}

The initializing behavior ($B_{INI}$) is used for initializing the genuine chunks in LTMs during the initialization stage ($t=0$). For each row in SPEC-MP, $B_{INI}$ executes the $R_{IE}$ instance to obtain $\$CH_{M}$ for $CH_{M} \in M_{A}$ of all agents and $CH_{M} \in M_{SG}$ of the interactive center (as the output types shown in Table \ref{Tab:iotypes}).
 
Duraing the runtime ($t>0$), the chunks in $M_{A}$ and $M_{SG}$ are respectively updated by the $M_{A}$-\emph{updating} behavior ($B_{UA}$) and $M_{SG}$-\emph{updating} behavior ($B_{USG}$). 

For each row of SPEC-MP, the input/output parameters $(CH_{(M)}, CH_{(U)})$ of $R_{UE}$ are linked to the corresponding cells in $M_{A}$ and $M_{BA}$ if $CH_{M}\in M_{A}$, or in $M_{SG}$ and $M_{BSG}$ if $CH_{M}\in M_{SG}$. Then in each cycle, each $B_{UA}$ or $B_{USG}$ executes the corresponding $R_{UE}$ instance if the corresponding buffer cell is not empty. 

\subsubsection{Collecting Behavior}
\label{sec:bupd}

The collecting behavior ($B_{CO}$) is used for managing the dependent chunks in $M_{SD}$ of the interactive center. At $t=0$, $B_{CO}$ forms each dependent chunk $CH_{M} = \{CH_{U(i)}|i\in[1, N]\}$ into $M_{SD}$, where $CH_{U(i)}$ is from $M_{A}$ of the $i$th agent. During the runtime, dependent chunks in $M_{SD}$ are automatically updated if the reference chunks in $M_{A}$ of any agents are updated.

\subsubsection{Submitting Behavior}
\label{sec:bsub}

The \emph{submitting} behavior ($B_{SUB}$) submits chunks into $M_{BA}$ and $M_{BSG}$, given $M_{A}$, $M_{G}$, and a updating list $E_{UPD}$ of genuine chunk identifiers. For each chunk identifier $CH_{M} \in E_{UPD}$, the corresponding row in SPEC-MP is found, and then a cloned chunk of $CH_{U} \in  M_{A} \cup M_{G}$ is submitted into the corresponding buffer cell in $M_{BA}$ or $M_{BSG}$ if $CH_{M} \in M_{A}$ or $CH_{M} \in M_{SG}$.

\subsubsection{Generating Behavior}
\label{sec:bgen}

The \emph{generating} behavior ($B_{GEN}$) generates a chunk with the solution property into $M_G$. Based on a given $ID_{G}$, the corresponding ESH in SPEC-G is chosen. Then $B_{GEN}$ executes the $R_{GE}$ instance and generates a chunk $CH_{OG}$ into $M_G$ by using the input chunk list $E_{IG} \in M_{A} \cup M_{S}$.

\subsubsection{Meta-Managing Behavior}
\label{sec:bmm}

The \emph{meta-managing} behavior ($B_{MM}$) picks the algorithm case with a given $ID_{C}$ from SPEC-MM. Afterward, one of the executive rows in the CGO case is probabilistically selected, according to the associated $C_{W}$ values. Afterward, $ID_{G}$ and $E_{UPD}$ in the selected executive row are used as the inputs for consecutively executing the $B_{GEN}$ and $B_{SUB}$ behaviors.

\subsection{CGO Framework: Execution Process}
\label{sec:exec}

Algorithm \ref{alg:cgoframework} gives the execution process of the CGO framework. In each line, the working module (entity), the required inputs, and the outputs or updated modules are provided. 

\begin{algorithm*} [htb]                     
\caption{The execution process of the CGO framework, given a script and a toolbox}   
\label{alg:cgoframework}                           
\begin{algorithmic}[1]                    
\REQUIRE The optimization problem $F_{P}$; A CGO case ($ID_{C}$ + script parameters: $N$, $T$, $\cdots$)
\STATE Facilitator: $\langle \text{SPEC-F}, F_{P}\rangle \rightarrow \langle F_{N}, F_{R}\rangle$~~~~~~~~~~~~~~~~~~~~~~~~~~~~~~~~~~~~~~~~~~~~~~~\COMMENT{$F_{R}$ is used by all agents and the interactive center}
\STATE $B_{INI}$: $\text{SPEC-MP} \rightarrow \langle \{M_{A(i)}|i\in[1, N]\}, M_{SG}\rangle$ ~~~~~~~~~~~~~~~~~~~~~~~~~~~~~~~~~~~~~~~~~~~~~~~~~~~~~~~~~~~~~~~~\COMMENT{$R_{IE}$ instances, Section \ref{sec:bini}}
\STATE $B_{CO}$: $\text{SPEC-MP} \rightarrow M_{SD}$~~~~~~~~~~~~~~~~~~~~~~~~~~~~~~~~~~~~~~~~~~~~~~~~~~~~~~~~~~~~~~~~~~~~~~~~~~~~~~~~~~~~~~~~~~~~~~~~~~~~~~~~~~~~~~~~~\COMMENT{Section \ref{sec:bini}}
\FOR{$t$ = 1 to $T$} 
\STATE Facilitator: $\langle F_{R}^{(t)},  M_{S} \rangle \rightarrow  F_{R}^{(t+1)}$~~~~~~~~~~~~~~~~~~~~~~~~~~~~~~~~~~~~~~~~~~~~~~~~~~~~~~~~~~~~~~~~~~~~\COMMENT{[Optional] landscape transformation}
\FOR{$i$ = 1 to $N$} 
\STATE $B_{MM(i)}$: $\langle ID_{C}, \text{SPEC-MM} \rangle \rightarrow \langle ID_{G}, E_{UPD} \rangle$~~~~~~~~~~~~~~~~~~~~~~~~~~~~~~~~~~~~~~~~~~~~~~~~~~~~~~~~~~~~~~~~~~~~~~~~~~~~~~~~~~~~~\COMMENT{Section \ref{sec:bmm}}
\STATE $B_{GEN(i)}$: $\langle ID_{G}, \text{SPEC-G}, M_{A(i)}, M_{S} \rangle \rightarrow CH_{OG} \in M_{G(i)}$~~~~~~~~~~~~~~~~~~~~~~~~~~~~~~~~~~~~~~~~\COMMENT{$R_{GE}$ instances, Section \ref{sec:bgen}}
\STATE $B_{SUB(i)}$: $\langle E_{UPD}, \text{SPEC-MP}, M_{A(i)}, CH_{OG} \rangle \rightarrow \langle M_{BA(i)}, M_{BSG}\rangle$~~~~~~~~~~~~~~~~~~~~~~~~~~~~~~~~~~~~~~~~~~~~~~~~~~~~~\COMMENT{Section \ref{sec:bsub}}
\STATE $B_{SK}$: $\langle CH_{OG}, F_{N} \rangle \rightarrow \vec{x}^{\bar{*}}$~~~~~~~~~~~~~~~~~~~~~~~~~~~~~~~~~~~~~~~~~~~~~~~~~~~~~~~~~~~~~~~~~~~~~~~~~~~~~~~~~~~~~~\COMMENT{Stores the best-so-far solution}
\ENDFOR
\STATE $B_{USG}$: $\langle \text{SPEC-MP}, M_{BSG}, M_{SG}^{(t)} \rangle \rightarrow M_{SG}^{(t+1)}$~~~~~~~~~~~~~~~~~~~~~~~~~~~~~~~~~~~~~~~~~~~~~~~~~~~~~~~~~~~~~~~~~~~\COMMENT{$R_{UE}$ instances, Section \ref{sec:bupd}}
\FOR{$i$ = 1 to $N$} 
\STATE $B_{UA(i)}$: $\langle \text{SPEC-MP}, M_{BA(i)}, M_{A(i)}^{(t)} \rangle \rightarrow$ $M_{A(i)}^{(t+1)}$~~~~~~~~~~~~~~~~~~~~~~~~~~~~~~~~~~~~~~~~~~~~~~~~~~~~~~~\COMMENT{$R_{UE}$ instances, Section \ref{sec:bupd}}
\ENDFOR
\STATE {\text{/*} $M_{SD}^{(t)} \rightarrow M_{SD}^{(t+1)}$ is automatically updated by the dependence built by $B_{CO}$ \text{*/}}
\ENDFOR
\RETURN $\vec{x}^{\bar{*}}$ ~~~~~~~~~~~~~~~~~~~~~~~~~~~~~~~~~~~~~~~~~~~~~~~~~~~~~~~~~~~~~~~~~~~~~~~~~~~~~~~~~~~~~~~~~~~~~~~~~~~~~~~~~~~~~~~\COMMENT{Returns the best-so-far solution}
\end{algorithmic}
\end{algorithm*}

In Line 1, $F_{P}$ is formulated into $F_{N}$ and  $F_{R}$ by forming the elements $<S_{P}, R_{M:N}, R_{M(R)}, AUX>$. In Lines 2 and 3, all long-term memories used by the agents and the interactive center are initialized by using $B_{INI}$ and $B_{CO}$.
After the initialization, the CGO framework runs in iterative learning cycles, in which each learning cycle $t\in [1, T]$ is executed between Lines 5--16.

In line 5, an option is provided for the facilitator for updating $F_{R}^{(t)}$ by using a chunk in $M_{S}$. In lines 7--10, each agent $i\in[1, N]$ is executed. In Line 7, $B_{MM(i)}$ is executed to select an executive row, which contains $ID_{G}$ and $E_{UPD}$, in SPEC-MM. In Line 8, the embedded search heuristic (ESH), which is named $ID_{G}$ in SPEC-G, is triggered to generate its output chunk $CH_{OG} \in M_{G(i)}$ by using the list of input chunks $E_{IG} \in  M_{A(i)}\cup M_{S}$. In Line 9, the buffer cells in $M_{BA(i)}$ and $M_{BSG}$, which are corresponding to the updating list $E_{UPD}$ in LTMs, are filled by 
$B_{SUB(i)}$ by using SPEC-MP. In Line 10, the solution contained in $CH_{OG}\in M_{G(i)}$ is processed by $B_{SK}$ to obtained the best-so-far solution $\vec{x}^{\bar{*}}$, based on the quality evaluation by using $F_{N}$. During Lines 6--11, all LTMs remain unchanged. In Line 12--15, $M_{SG}$ and each $M_{A(i)}$ are independently updated by $B_{USG}$ and each $B_{UA(i)}$. Line 16 mentions the fact that $M_{SG}$ is automatically updated if the corresponding chunks in $M_{A}$ of agents are updated. Finally, $\vec{x}^{\bar{*}}$ is returned while the framework is terminated.

\subsection{System Characteristics}
\label{sec:Characteristics}

The CGO system has three characteristics: (1) the CGO framework can support a cooperative group; (2) each agent holds a customized portfolio of ESHs; and (3) The framework is driven by a multilayer script working on knowledge components in the CGO toolbox. 

\subsubsection{Cooperative Group}
\label{sec:cgroup}

In principle, the CGO framework can support three kinds of groups: (1) a \emph{nominal group}, in which each agent performs lifetime learning by only using its individual memory ($M_{A}$); (2) a \emph{stigmergic/evolutionary group}, in which each agent does not possess its $M_{A}$, but agents can indirect cooperate with their peers through the social memory ($M_{S}$); and (3) a \emph{cooperative group}, in which each agent performs a mix of the individual and social learning by using both $M_{A}$ and $M_{S}$. Both nominal and cooperative groups have been used for studying group creativity \cite{Goncalo:2006p1071,Paulus:2000p1114}.

The cooperative group is an advanced algorithm designed by natural evolution over millions of years. The paradigm helps striking a natural balance between exploitation and exploration in the problem landscape. In a cooperative group, the agents explore in a parallel way with their individual memory, as well as cooperate with their peers through the group memory.

For each agent, its individual memory \cite{Glenberg:1997p1390,Ericsson:1995p1364}, i.e., $M_{A}$, supports its lifetime learning \cite{Curran:2006p1143}, e.g., ``trial-and-error'', for discovering novel knowledge based on experience. 
A sequence of the chunks (or thoughts \cite{Ericsson:1995p1364}) updated in the same cell can be regarded as a ``trajectory'' \cite{Glenberg:1997p1390} along with learning cycles. 
An algorithmic example of individual learning strategies is stochastic local search \cite{Hoos2004}. In a group, individual learning is essential for social learning to be useful \cite{Laland:2004p1076}, by escaping from some maladaptive outcomes \cite{Boyd2011}.

For a group of agents, $M_{S}$ might be referred as public memory \cite{Danchin:2004p1204} or group memory \cite{Dennis:1993p1298,Satzinger1999}. $M_{SD}$ contains non-private chunks that can be observed from $M_{A}$ of the agents, whereas $M_{SG}$ contains all genuine chunks that are not possessed by any agents, e.g., pheromone trails in an ant colony \cite{Socha2008} and the group memory in \emph{brainstorming}  \cite{Dennis:1993p1298}. Many animals and human beings are able to learn socially by observing their peers and/or utilizing external knowledge \cite{Laland:2004p1076,Danchin:2004p1204}. 

Thus, each agent possesses a mixed cultural learning capability \cite{Curran:2006p1143,Tomasello:1993p1330,Galef:1995p1128,Boyd2011} that operates with both the individual and social memories. The social memory contains gradually accumulated and recombined adaptive knowledge \cite{Boyd2011} for accelerating the learning process, whereas the individual memory preserve some promising minority patterns \cite{Nemeth:1986p980} for supporting the capability of escaping from some maladaptive outcomes \cite{Boyd2011}, which is essential for the social memory to be useful \cite{Laland:2004p1076}. The emergence of solutions in the group level might also share some essences with collective intelligence \cite{Woolley2010}.

In each cycle, the agents might be different in not only the chunks in their $M_{A}$ but also the executive rows picked by their $B_{MM}$. Even a nominal group becomes a portfolio of heterogeneous algorithms \cite{Huberman:1997p1159,Streeter:2008p1072}, which may achieve better overall performance by exploiting the large variance among the performance of the agents.

Allowing for cooperation among the agents may improve search performance \cite{Huberman:1997p1159} by enabling agents to circumvent their own cognitive limitations. Compared to a nominal group, the interaction may enhance the group creativity \cite{Paulus:2000p1114}, as shown in \emph{brainstorming} \cite{Dennis:1993p1298,Kohn2011a}.

Compared to a stigmergic group, a cooperative group has two major features due to the possession of personal memories by the agents. First, the agents may explore in a parallel way, while the diversity of positive patterns, even those in minority \cite{Nemeth:1986p980}, can be preserved in a more reliable way. Individualism in a group may foster the group creativity by encouraging uniqueness \cite{Goncalo:2006p1071}. Second, the cooperative mechanism in a group needs not to be designed very carefully since the public information does not have an overwhelming impact on accumulated knowledge in the system. 

For the viewpoint of \emph{population-based algorithms}, stigmergic groups, e.g., ES \cite{Runarsson:2005p1196}, GA \cite{Deb:2000p1200}, MA \cite{Ong2006}, ACO \cite{Socha2008}, and CA \cite{Reynolds:2008p1078}, are commonly studied, where $M_{S}$ might contains different chunks, e.g., an evolutionary population \cite{Runarsson:2005p1196,Ong2006}, a pheromone matrix \cite{Socha2008}, and external belief \cite{Reynolds:2008p1078}. A nominal group can be seen as a portfolio \cite{Huberman:1997p1159,Streeter:2008p1072} of independent local search agents \cite{Hoos2004}. 

Independent local search agents can only use blind disturbance \cite{Hoos2004}, whereas cooperative agents can be guided by adaptive clues in the social memory, when they are trying to escape from some local valleys in the problem landscape. In a stigmergic group such as GA, the population diversity must be explicitly maintained by frequently applying blind disturbance, e.g., mutation operators, whereas a cooperative group keeps novel and diverse states in individual memory of the agents.

\subsubsection{Algorithm Portfolios}
\label{sec:toolbox}

In the CGO system, each agent holds a portfolio \cite{Huberman:1997p1159} of embedded search heuristics (ESHs). The meta-management behavior can be viewed as a task-switching schedule \cite{Streeter:2008p1072} to interleave the execution of a portfolio \cite{Huberman:1997p1159} of ESHs across learning cycles. Any executive rows using the nodes of the same tree in the updatable graph may be cooperative by default, or they can be turned into cooperative rows by using the customized mode, if necessarily. The cooperation among ESHs may further improve performance \cite{Huberman:1997p1159}. The cooperative algorithm portfolio is strengthened in the cooperate group, since  novel cooperation results can be easily diffused though the group memory, and detrimental results might be isolated in individual memory of agents. 

From a user-oriented perspective, the practical problem sets are different during different periods for different users. According to the No Free Lunch (NFL) theorems \cite{Wolpert:1997p1149}, it is impossible to obtain an omnipotent algorithm case for a sufficiently diverse set of problem instances. Thus, it is rational to tackle the problem set faced by each user during a sufficiently long period, by using a portfolio of fast-and-frugal heuristics. New heuristics, which mainly tackle some of new problems unsolved by existing heuristics, can be implemented into the portfolio over time.

According to its $E_{(IG)}$, each ESH might possess one of the four search properties:
a) \emph{scratch search}, which has $E_{(IG)}=\varnothing$; b)
\emph{individual learning}, which only uses the chunks in
$M_A$; b) \emph{social learning}, which only uses the chunks
in $M_{SG}$; and d) \emph{cultural learning} (or \emph{socially-biased individual learning} \cite{Galef:1995p1128}), which employs the input chunks in both $M_A$ and $M_S$. In principle, the agents in a cooperative group might use mixed strategies, as long as cultural learning strategies play a nontrivial role. For example, in GSO \cite{He:2009p1043}, only scroungers use a cultural learning strategy, whereas producers and rangers employ individual learning strategies.

There is a basic paradigm shift in supporting the algorithm space. Traditional methods mainly use the space of setting parameters for tuning/controlling the algorithm performance \cite{Eiben:1999p1133}. Each algorithm with setting parameters might support a huge algorithm space, but only a few algorithm cases are competent for some problem instances. The number of useful algorithm cases might be dropped to much less if the overlap among the performance of different algorithm cases is considered. The CGO algorithm space is mainly defined upon a portfolio of basis ESHs, in which each ESH is competent, which captures explicit/implicit domain-specific features, for some problem instances. Furthermore, CGO cases can be defined in an independent or cooperative way to stretch for solving most problem instances. Moreover, allowing heterogeneous inputs/outputs increases the chance of finding competent ESHs. The total portfolio size can be maintained to be small by adding new ESHs that are competent for new problems as well as removing obsolete ESHs. 

\subsubsection{Multilayer Script}

The CGO system is a development framework that is driven by an multilayer script. This is essential for supporting the vision of the adaptive box, since implementing many stand-alone search heuristics might require quite an effort, let alone flexibly supporting the cooperative search among heterogeneous search heuristics that are sharing customized memory elements. 

In the upper three layers of the CGO script, each layer contains some elemental rows. A new row can be added into a layer to provide more choices for implementing new rows into higher layers; and an old row can be removed from a layer if it is not used by any higher layers. Thus, any competent knowledge components can be easily accumulated, whereas obsolete knowledge component can be easily removed, without leading to any risk to interfere existing algorithm cases.
 
Each layer might provide nontrivial knowledge for the higher layer. SPEC-MP forms a simple memory protocol ontology for the interaction between the agents and the interactive center. The corresponding updatable graph is not only useful for checking the validity of SPEC-MP, but also for defining feasible input/output chunks for each ESH. SPEC-MP also provides a nontrivial support for the stability of cooperative search among different ESHs. The actual performance of each ESH in SPEC-G provides essential knowledge for designing promising portfolios \cite{Streeter:2008p1072,Huberman:1997p1159} in SPEC-MM.   

The implementation process may mainly occur in higher levels, once there are enough supports from lower layers. Eventually, almost all operations will take place in SPEC-MM for finding suitable portfolios, either cooperative or not, if a sufficiently large number of ESHs are implemented.
 
\section{Implementation for Constrained Optimization}
\label{sec:COP}

For each problem $F_{P}$ of a specific type, the CGO system can be concretely implemented. Here the constrained optimization problem is used for demonstrating the implementation process. We first introduce the problem, then describe problem-specific algorithmic components. Here we provide full details of these components for easily reproducing the algorithms, but it might be worthy to keep in mind that each component is a binary object with specific input/output parameters for end users.  

\subsection{The Constrained Optimization Problem}

THe $F_{P}$ of the constrained optimization problem can be defined as follows \cite{Deb:2000p1200}:
\begin{equation}
\left\{\begin{array}{l} {{\rm Minimize}:~~f(\vec{x}){\rm \; \; } } \\ {{\rm Subject\; to:\; \; }g_{j} (\vec{x})\in [\underline{c}_{j} ,\bar{c}_{j} ]_{{\mathbb R}} {\rm \; \; \; \; \; \; \; \; \; \; }(j\in [1,J]_{{\mathbb Z}} )} \end{array}\right.
\end{equation}
in which $\vec{x}=(x_{[1]} ,\cdots,x_{[D]} ) \in S_{P} \subset {\mathbb R}^{D}$ is a state within the space $S_{P}$ which is a \emph{D}-dimensional Euclidean space with the \emph{boundary constraints} $x_{[d]} \in [\underline{x}_{[d]} ,\bar{x}_{[d]} ]_{{\mathbb R}} $ for $\forall d\in [1,D]_{{\mathbb Z}} $, $f(\vec{x})$ is the \emph{objective function}, and each $g_{j} (\vec{x})$ is a \emph{constraint function} with two constant boundary values $\underline{c}_{j} $ and $\bar{c}_{j} $ ($\underline{c}_{j} \le \bar{c}_{j} $). The feasible space $S_{PF}$ is defined as $S_{PF}$=\{$\vec{x}$\textbar $g_{j} (\vec{x})\in [\underline{c}_{j} ,\bar{c}_{j} ]_{{\mathbb R}} ,{\rm \; }\forall j\in [1,J]_{{\mathbb Z}} $; $\vec{x}\in S_{P}$\}. Any solutions in $S_{PO}$ are located in $S_{PF}$, by treating all constraints as hard constraints. 
We define $\underline{\bar{c}}_{\min}=\operatorname{min}_{j=1}^{J}(\bar{c}_{j} -\underline{c}_{j})$ for convenience. If $\underline{c}_{j} \equiv \bar{c}_{j}$, the $j$th constraint is called an \emph{equality constraint}, which is preprocessed into a relaxed form by using $g_{j} (\vec{x})\in [\underline{c}_{j} -\varepsilon _{H} ,\bar{c}_{j} +\varepsilon _{H}]$ with a tolerance parameter $\epsilon_{H}>0$.

\subsection{Quality Measurement}
\label{sec:COP_RQM}

The quality measurement is a macro subtype, i.e., $R_{M}$ = $<R_{QE}^{V2}, R_{QC}>$, which is realized as
\begin{equation}
{R}_{M}(\vec{x}_{(a)}, \vec{x}_{(b)})=R_{QC}(R_{QE}^{V2}(\vec{x}_{(a)},R_{QE}^{V2}(\vec{x}_{(b)})) 
\label{eq:QM}
\end{equation}
in which the \emph{quality-encoding} ($R_{QE}^{V2}$) rule is a background rule that calculates the intermediate quality of the only input state into a commonly-used data structure ($v_{_{CON}},v_{_{OBJ}}$), then the \emph{quality-comparing} rule ($R_{QC}$) evaluates any two ($v_{_{CON}}, v_{_{OBJ}}$) instances and returns TRUE or FALSE.

For $R_{QE}^{V2}(\vec{x})$, the output element $v_{_{OBJ}}$ is equal to the objective function value $f(\vec{x})$, and the output element $v_{_{CON}}$ is the summarized constraint violation value, i.e.,

\begin{equation}
v_{_{CON}}=\sum _{j=1}^{J}\left\{\begin{array}{l} {0} \\ {\underline{c}_{j} -g_{j} (\vec{x})} \\ {g_{j} (\vec{x})-\bar{c}_{j} } \end{array}\right. \begin{array}{l} {{\rm \; \; \; \; IF\; }g_{j} (\vec{x})\in [\underline{c}_{j} ,\bar{c}_{j} ]_{{\mathbb R}} } \\ {{\rm \; \; \; \; IF\; }g_{j} (\vec{x})<\underline{c}_{j} } \\ {{\rm \; \; \; \; IF\; }g_{j} (\vec{x})>\bar{c}_{j} } \end{array}
\end{equation}

Thus, the minimum value of $v_{_{CON}}$ is 0. For $\forall \vec{x}$, if there is $v_{_{CON}}$ $\equiv $0, then it means $\vec{x}\in S_{PF}$.

\subsubsection{Existing Quality Comparison Rules}
\label{sec:QCRules}

A basic usage is that various existing constraint-handling techniques may be realized by using different $R_{QC}$ rules.

A $R_{QC}$ rule ($R_{QC}^{O}$) returns TRUE, if there is: a) $v_{_{CON}(a)} < v_{_{CON}(b)}$; or b) $v_{_{CON}(a)} \equiv v_{_{CON}(b)}$ and $v_{_{OBJ}(a)} $ $\le v_{_{OBJ}(b)}$. The $R_{QC}^{O}$ rule satisfies the following criteria \cite{Deb:2000p1200}: a) $\vec{x}_{(a)} \in S_{PF}$ is preferred to $\vec{x}_{(b)} \notin S_{PF}$; b) between two states within $S_{PF}$, the one having a smaller objective function value is preferred; c) between two states out of $S_{PF}$, the one having a smaller constraint violation is preferred. It has been widely used in some existing work \cite{Zhang:2003p1404,MezuraMontes:2005p1325,Lu:2008p1507}.

The \emph{penalized} $R_{QC}$ rule ($R_{QC}^{P}$) returns TRUE, if there is $v_{all(a)} \le v_{all(b)} $, in which $v_{all} =v_{_{OBJ}} +C_{AP} \cdot v_{_{CON}} $ and $C_{AP}\ge 0$ is a static penalty coefficient. The static penalty term has been used as a popular technique \cite{Deb:2000p1200}. However, deciding a good penalty coefficient for each specific problem instance might be a rather difficult optimization problem itself.

The \emph{stochastic} $R_{QC}$ rule ($R_{QC}^{S}$) returns TRUE, if there is: a) $v_{_{CON}(a)} <v_{_{CON}(b)}$; or b) $v_{_{OBJ}(a)} \le v_{_{OBJ}(b)}$ as $v_{_{CON}(a)} \equiv v_{_{CON}(b)} $ or $U_{{\mathbb R}} <C_{PF}$, in which $C_{PF}\in [0,1]_{{\mathbb R}} $ is a setting parameter. This rule is used in the stochastic ranking technique \cite{Runarsson:2005p1196}.

The \emph{static-relaxing} $R_{QC}$ rule ($R_{QC}^{RS}$) returns TRUE, if there is: a) $v_{_{CON}(a)} <v_{_{CON}(b)}$ as $v_{_{CON}(b)} >C_{ER}$; or b) $v_{_{OBJ}(a)} \le v_{_{OBJ}(b)} $ as both $v_{_{CON}(a)} ,v_{_{CON}(b)} \le C_{ER} $. Here $C_{ER}\ge $0 is a relaxing value.

A \emph{dynamic-relaxing} $R_{QC}$ rule is then defined as a tuple $<R_{QC}^{RS}$, $R_{ADJ}>$, where the \emph{adjusting} rule ($R_{ADJ}$) dynamically adjusts the $C_{ER}$ value of $R_{QC}^{RS}$ \cite{Hamida:2002p1205,Xie:2004p1408}. 

Among these $R_{QC}$ rules, only $R_{QC}^{O}$ leads to a natural landscape. However, the $S_{PF}$ of a natural landscape is critically shaped by the boundary values of constraint functions. If the $\underline{\bar{c}}_{\min}$ value of $F_{P}$ is not large enough, $S_{PF}$ may be long and narrow valleys, which can be divided into multiple segments of the ridge function class \cite{Beyer:2001p1308} of unknown directions. Searching in such valleys is very challenging since improvement intervals \cite{Salomon:1996p1280} toward better solutions are predominantly too small.

For a landscape with the $R_{QC}^{RS} $ rule, its quasi-feasible space is $S_{PF}^{'} =\{ \vec{x} | v_{_{CON}} (\vec{x})\le C_{ER} ;\vec{x}\in S_{P} \} \supseteq S_{PF}$. A good attribute is that $S_{PF}^{'}$ always shrinks when $C_{ER}$ decreases, till $S_{PF}^{'} \rightarrow S_{PF}$ for $C_{ER} \rightarrow 0$. To increase improvement intervals and to approach the optimum from both feasible and infeasible space, it is rational to sufficiently relax $S_{PF}^{'}$ at the early stage and gradually shrinks $S_{PF}^{'}$ to $S_{PF}$, by adjusting the $C_{ER}$ value of $R_{QC}^{RS} $ from large to small, during the run-time. Some dynamic adjustment techniques have been proposed \cite{Hamida:2002p1205,Xie:2004p1408}. The basic experience is that the adjustment pace is a key issue for the performance. In the next section (Section \ref{sec:RRAdj}), we will consider a slightly modified adjusting rule for maintaining a controllable adjustment pace. 

Some of these $R_{QC}$ rules, including  $R_{QC}^{O}$, $R_{QC}^{P}$, $R_{QC}^{S}$, and the dynamic adjustment method in \cite{Hamida:2002p1205} have been used in existing algorithms discussed in Section \ref{sec:ExsitAlgs}. 

\subsubsection{Adaptive Ratio-Reaching Adjustment}
\label{sec:RRAdj}

The \emph{ratio-reaching} $R_{ADJ}$ rule ($R_{ADJ}^{RR}$) adjusts $C_{ER}^{(t)} $ by using the $\bar{\underline{c}}_{\min } $ value of $F_{P}$ and the maximum number of cycles ($T$). Furthermore, it has four setting parameters: $C_{RRE}$, $C_{RNU}$, $C_{RTU}$, and an input state set called $\$\vec{x}_{FB}$. For convenience, we define $t_{TH} ={\rm INT}(C_{RTU} \cdot T )$, in which $C_{RTU} \in [0,1]_{{\mathbb R}} $ and the function INT(\emph{t}) returns the closest integer value of its input \emph{t}. 

There are $C_{ER}^{(0)}=0$ and $C_{ER}^{(1)}$ is set as the maximum $v_{_{CON}} $value \cite{Xie:2004p1408} of all the states in $\$\vec{x}_{FB}$. As $t>t_{TH}$, there is $C_{ER}^{(t)} $=0 \cite{Xie:2004p1408}. Thus there are totally ($T-t_{TH}$) learning cycles left for fulfilling the final search process in $S_{PF}$. For $1\le t\le t_{TH}$, if $c_{RNC}^{(t)} >C_{RNU}$, there is
\begin{equation} 
C_{ER}^{(t+1)} =C_{ER}^{(t)} \cdot \left({c_{ERE} \mathord{\left/ {\vphantom {c_{ERE}  C_{ER}^{(t)} }} \right. \kern-\nulldelimiterspace} C_{ER}^{(t)} } \right)^{{\rm 1/(}t_{TH} -t+1{\rm )}}
\label{eq:ERE}
\end{equation}
in which $C_{RNU}$ $\in [0,1]_{{\mathbb R}}$, $c_{RNC}^{(t)} $ is the ratio of the states within the current $S_{PF}^{'} $ over all the states in $\$\vec{x}_{FB}$ \cite{Hamida:2002p1205}, $c_{ERE} =C_{RRE} \cdot \bar{\underline{c}}_{\min } /2$ is an expected value of $C_{ER}^{(t_{TH})}$, and $C_{RRE}$ is a positive constant.

Compared to the previous methods \cite{Hamida:2002p1205,Xie:2004p1408}, the minor modification in Eq. \ref{eq:ERE} aims in keeping a stable $c_{RNC}^{(t)} $ value, while maintaining an adaptive pace for adjusting $C_{ER}^{(t)}$ to the expected $c_{ERE}$ value at $t = t_{TH}$.

\subsubsection{Incorporate Global Knowledge}
\label{sec:COP_O3R}

The equality constraints can be seen as a typical problem feature that is known in advance. The $R_{QC}^{O3R}$ rule is a macro $R_{QC}$ rule integrating $R_{QC}^{O} $ and $R_{QC}^{RRR} $=$<R_{QC}^{RS} $, $R_{ADJ}^{RR} >$ by a simple policy, i.e., $R_{QC}^{RRR} $ is executed if $\bar{\underline{c}}_{\min} \le 2\cdot \varepsilon _{H} $, otherwise $R_{QC}^{O} $ is executed. Thus problem instances with and without equality constraints are tackled by the $R_{QC}^{RRR}$ and $R_{QC}^{O} $ rules, respectively.

\subsection{The Facilitator}
\label{sec:COP_FAC}

Based on $F_P$, the facilitator is implemented from a tuple, i.e., $<S_{P}, R_{M:N}, R_{M(R)}, AUX>$, where the $R_{M:N}$ rule in $F_{N}$ and the $R_{M(R)}$ rule in $F_{R}$ are respectively realized as $<R_{QE}^{V2}, R_{QC}^{O}>$ and $<R_{QE}^{V2}, R_{QC(R)}>$ (based on Eq. \ref{eq:QM}), and $AUX$ only contains the \emph{D}-dimensional Euclidean space $S_{P}$ defined by the boundary constraints. 

Here $AUX$ is represented in a concise form. Following a practical setting of black-box optimization, no function details and gradient information are considered. The detail information of function values for candidate states and the boundary values for all constraint functions are encapsulated in the $R_{M(R)}$ rule.

For the facilitator, only $R_{QC(R)}$ can be assigned, but a high flexibility is still retained since $R_{QC(R)}$ can be realized in various forms (e.g., different $R_{QC}$ rules defined in Sections \ref{sec:QCRules} and \ref{sec:COP_O3R}).

\subsection{Elemental Generating Rules}
\label{sec:COP_RGE}

Three $R_{GE}$ rules are extracted from existing algorithms. In addition, boundary-handling methods are integrated into $R_{GE}$ rules according to available knowledge. 

\subsubsection{Differential Evolution $R_{GE}$ Rule }

The \emph{differential-evolution} $R_{GE}$ rule ($R_{GE}^{DE} $) is extracted from differential evolution (DE) \cite{Price:2005p2148}. Its $E_{(IG)}$ has two chunks, i.e., \{$\vec{x}_{P} $, $\$\vec{x}_{DP} $\}. Its setting parameters include $C_{F}$, $C_{CR}$, and $C_{CG}$, in which $C_{CR} \in [0,1]_{{\mathbb R}}$ is the crossover constant, $C_{F} >0$ is the scale constant. 

The $R_{GE}^{DE} $ rule generates one state $\vec{x}_{C}$ as its $CH_{(OG)}$ by using the following steps:

a) Create a list of states, i.e., \{$\vec{x}_{(a)} $,$\vec{x}_{(b)} $,$\vec{x}_{(c)} $,$\vec{x}_{(d)} $\}, by selecting from $\$\vec{x}_{DP} $ at random;

b) Obtain $c_{DR}$, which is randomly chosen from $[1,D]$;

c) For the $d$th dimension, if $U_{{\mathbb R}} <C_{CR}$ or $d\equiv c_{DR}$,
\begin{equation}
\begin{array}{l} 
{x_{C[d]} = x_{(P)[d]} +C_{CG} \cdot (x_{(g)[d]} -x_{(P)[d]})}\\
{\rm \; \; \; \; \; \; \; \; \; \; }{+C_{F} \cdot (x_{(a)[d]} -x_{(b)[d]} +x_{(c)[d]} -x_{(d)[d]} )}
\end{array}
\end{equation}
in which $\vec{x}_{(g)} = R_{SEL}^{G}(\$\vec{x}_{DP})$ (defined in Section \ref{sec:bgRules}) is the state with the best quality in $\$\vec{x}_{DP}$, and $C_{CG} \in [0, 1]$ is the ratio between  $x_{P[d]}$ and $x_{(g)[d]}$. Here $\vec{x}_{C}$ is respectively generated around $\vec{x}_{(g)}$ and $\vec{x}_{P}$, if $C_{CG}$ is assigned as 1 and 0. 

Its boundary-handling method is realized in a simple way: for the $d$th dimension, $x_{C[d]}$ is randomly chosen from $[\underline{x}_{[d]} ,{\rm \; }\bar{x}_{[d]} ]_{{\mathbb R}} $ if there is $x_{C[d]} \notin [\underline{x}_{[d]} ,{\rm \; }\bar{x}_{[d]} ]_{{\mathbb R}} $.

\subsubsection{Particle Swarm $R_{GE}$ Rule}

The \emph{particle swarm} $R_{GE}$ rule ($R_{GE}^{PS} $) is extracted from the operation of each particle in PSO \cite{Kennedy2001}.  Its $E_{(IG)}$ possesses four input chunks, i.e. \{$\vec{x}_{O} $,$\vec{x}_{R} $,$\vec{x}_{P} $, $\$\vec{x}_{DP}$\}, and its $CH_{(OG)}$ is $\vec{x}_{C}$. Furthermore, its setting parameters include $C_{A}>2$ and $C_{B}>2$.

For the $d$th dimension, $\vec{x}_{C} $ is generated as

\begin{equation}
\begin{array}{l} {x_{C[d]} =x_{R[d]} +c_{K} \cdot {\rm DIS}(x_{R[d]} ,x_{O[d]} ,d)}\\{\rm \; \; \; \; \; \; \; \; \; \; }{+C_{A} \cdot U_{{\mathbb R}} \cdot {\rm DIS}(x_{P[d]} ,x_{R[d]} ,d)} \\ {\rm \; \; \; \; \; \; \; \; \; \; }{+C_{B} \cdot U_{{\mathbb R}} \cdot {\rm DIS}(x_{(g)[d]} ,x_{R[d]} ,d){\rm \; }} \end{array}
\end{equation}
in which $c_{K} ={2\mathord{\left/ {\vphantom {2 (}} \right. \kern-\nulldelimiterspace} (} \sqrt{\varphi \cdot (\varphi -4)} +\varphi -2)$, $\varphi$=$C_{A}$+$C_{B}>4$, $\vec{x}_{(g)}$ is the state with the best quality in $\$\vec{x}_{DP}$, $U_{{\mathbb R}}$ is a real value randomly selected in [0, 1]$_{{\mathbb R}} $, and DIS($x_{(a)}$, $x_{(b)}$, $d$) calculates the distance between $x_{(a)}$ and $x_{(b)}$ at the $d$th dimension of $S_{P}$ \cite{Xie:2005p1406}, i.e.,
\begin{equation}
{\rm DIS}(x_{(a)} ,x_{(b)} ,d)=\left\{\begin{array}{l} {\underline{\bar{x}}_{[d]} +y {\rm \;\;  IF\; }y <-\underline{\bar{x}}_{[d]} /2} \\ {\underline{\bar{x}}_{[d]} -y {\rm \;\;  IF\; }y >\underline{\bar{x}}_{[d]} /2} \\ {y {\rm \; \; \; \; \; \; \; \; \; \; \; \; OTHERWISE}} \end{array}\right.
\end{equation}
in which  $\underline{\bar{x}}_{[d]} =\bar{x}_{[d]} -\underline{x}_{[d]} $ and $y =x_{(a)} -x_{(b)} $.

Finally, $\vec{x}_{C[d]}$ is repaired for $\forall \vec{x}_{C} \notin S_{P}$ \cite{Xie:2005p1406}, i.e., 

\begin{equation}
x_{C[d]}=\left\{\begin{array}{l} {\bar{x}_{[d]} -(\underline{x}_{[d]} -x_{C[d]})\% \underline{\bar{x}}_{[d]} {\rm \;\;  IF}{\rm \; }x_{C[d]}<\underline{x}_{[d]}} \\ {\underline{x}_{[d]} +(x_{C[d]}^{} -\bar{x}_{[d]} )\% \underline{\bar{x}}_{[d]} {\rm \;\;  IF}{\rm \; }x_{C[d]}>\bar{x}_{[d]}} \end{array}\right.
\end{equation}

\subsubsection{Social Cognitive $R_{GE}$ Rule}

The \emph{social cognitive} $R_{GE}$ rule ($R_{GE}^{SC}$) is extracted from the operation of an agent in social cognitive optimization (SCO) \cite{Xie:2002p1415}. Its $E_{(IG)}$ possesses two inputs chunks, i.e., \{$\vec{x}_{R} $, $\$\vec{x}_{GR}$\}, and one output chunk, i.e., $\vec{x}_{C} $. Furthermore, $R_{GE}^{SC} $ has only one setting parameter of the integer type, i.e., $C_{NTB}>$0. The basic idea is to learn from a good model state in a public knowledge pool. 

The $R_{GE}^{SC} $ rule produces $\vec{x}_{C}$ using the following steps: 

a) Select a model state $\vec{x}_{(m)}$ from $\$\vec{x}_{GR}$, by using the $R_{SEL}^{TS}$ rule (as defined in Section \ref{sec:bgRules}) with the setting parameter values $C_{NTS}$=$C_{NTB}$ and $C_{BQ}$=TRUE;

b) Determine two states $\vec{x}_{(b)}$ and $\vec{x}_{(r)} $ from $\vec{x}_{R} $ and $\vec{x}_{(m)} $: If $R_{M(R)} (\vec{x}_{(m)} ,\vec{x}_{R} )\equiv $TRUE, then $\vec{x}_{(b)} $=$\vec{x}_{(m)} $ and $\vec{x}_{(r)} $=$\vec{x}_{R} $, otherwise $\vec{x}_{(b)} $=$\vec{x}_{R} $ and $\vec{x}_{(r)}$=$\vec{x}_{(m)}$;

c) Obtain the virtual promising space, called $S_{pv}$, which takes $\vec{x}_{(b)} $ as the center, and uses $\vec{x}_{(r)} $ to determine its range. For the $d$th dimension of $\vec{x} \in S_{pv}$, and $x_{(br)[d]} =|x_{(b)[d]} -x_{(r)[d]}|$, there is,
\begin{equation}
x_{[d]} \in [x_{(b)[d]} -x_{(br)[d]},{\rm \; }x_{(b)[d]} +x_{(br)[d]} ]_{{\mathbb R}}; 
\end{equation}

d) Generate $\vec{x}_{C}$ within $S_{pv}\cap S_{P}$ at random. 

\subsection{Implementation of CGO Script}

The CGO script is implemented over algorithmic components that are defined in previous sections: Some generic components are defined in Section \ref{sec:CGOLib}, whereas problem-specific components are defined in Sections \ref{sec:COP_RQM} and \ref{sec:COP_RGE}. Each setting parameter of a component instance is fixed, unless it is specially assigned as an overall script parameter. For simplicity, the script is shown in tables, but it can be easily converted into a standard format, e.g., extensible markup language (XML).

For the facilitator (realized in Section \ref{sec:COP_FAC}), SPEC-F is simply defined by assigning an $R_{QC}^{O3R}$ instance  (realized in Section \ref{sec:COP_O3R}) as $R_{QC(R)}$ in  its $R_{M(R)}$.
For the $R_{ADJ}^{RR}$ instance in $R_{QC}^{RRR}$, its parameters include $C_{RRE}$=10, $C_{RNU}$=0.5, $C_{RTU}$=0.5, and $\$\vec{x}_{FB}=\$\vec{x}_{DP} \in M_{S}$ (defined later in SPEC-MP).

Table \ref{tab:SPEC-MP} lists the elemental rows in SPEC-MP, for $M_{A}$=\{$\vec{x}_{O} $, $\vec{x}_{R}$, $\vec{x}_{P} $\}, $M_{SG}$=\{$\$\vec{x}_{GR} $\},  $M_{SD}$=\{$\$\vec{x}_{DP}$\}, and $M_{G}$=\{$\vec{x}_{C}$\}. Here the primary chunk interfaces include $\vec{x}$ and  $\$\vec{x}$. For $|\$\vec{x}_{GR}|$, the number of states is $C_{NGR}=4  \cdot N$. All genuine chunks are initialized by $R_{IE:X}^{RND}$. The dependent chunk is $\$\vec{x}_{DP}$=\{$\vec{x}_{P(i)} $\textbar $i\in [1,N ]$\}, in which $\vec{x}_{P(i)}$ is in $M_{A}$ of the $i$th agent. 
For illustrative purposes only, the corresponding updatable graph, which contains a single tree, is also shown in Table \ref{tab:SPEC-MP}.

As shown in Table \ref{tab:SPEC-G}, SPEC-G contains four embedded search heuristics (ESHs), in which each ESH has a $ID_{G}$, an $R_{GE}$ instance with setting parameter values, and the lists of input/output parameters $E_{IG}$ and $CH_{OG}$. Both G.DE1 and G.DE2 use $R_{GE}^{DE} $ instances, while G.PS and G.SC uses $R_{GE}^{PS} $ and $R_{GE}^{SC} $ instances, respectively. Besides, G.SC uses one chunk in $M_{SG}$, while the others uses one chunk in $M_{SD}$. For each ESH, its $E_{IG}$ contains the nodes in a sub-graph of the updatable graph in Table \ref{tab:SPEC-MP}. All the four ESHs use both $M_{A}$ and $M_{S}$ elements in their $E_{IG}$. For illustrative purposes only, the lists of all active genuine chunks, i.e., $E_{IGG}$, is also shown in Table \ref{tab:SPEC-G}. 

\begin{table*} [!ht]
\renewcommand{\arraystretch}{1.2}
\centering \caption{The memory protocol rows in SPEC-MP and the corresponding updatable graph}
\begin{tabular}{|c|c|c|c|c||c|} \hline 
$ID_{M}$ & $CH_{M}$ & $R_{IE}$ Instance & $R_{UE}$ Instance & $CH_{U}$ & Updatable Graph 
\\ \hline 
$M_{A}$ & $\vec{x}_{O}$ & \small $R_{IE:X}^{RND}$ & \small $R_{UE:S}^{D} $ & $\vec{x}_{R}$  & \multirow{5}{*}{\includegraphics[scale=1.0]{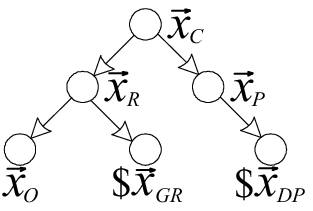}}\\ \cline{1-5}
 $M_{A}$ & $\vec{x}_{R}$ & \small $R_{IE:X}^{RND}$ & \small $R_{UE:S}^{D} $ & $\vec{x}_{C}$ & \\ \cline{1-5} 
 $M_{A}$ & $\vec{x}_{P}$ & \small $R_{IE:X}^{RND}$ & \small $R_{UE:S}^{G} $ & $\vec{x}_{C}$ &  \\ \cline{1-5}
 $M_{SG}$ & $\$\vec{x}_{GR}: C_{NGR}=4\cdot N$ & \small $R_{IE:X}^{RND}$ & \small $R_{UE:X}^{TS} :C_{NTW} =4$ & $\vec{x}_{R} $  & \\ \cline{1-5}
 $M_{SD}$ & $\$\vec{x}_{DP}$ & - & - & $\vec{x}_{P} $  & \\ \hline 
\end{tabular}
\label{tab:SPEC-MP}
\end{table*}

\begin{table*} [!ht]
\renewcommand{\arraystretch}{1.2}
\centering \caption{The generative rows in SPEC-G and the corresponding $E_{IGG}$ lists}
\begin{tabular}{|c|c|c|c||c|} \hline 
$ID_{G}$ & $R_{GE}$ Instance & $E_{IG}$ & $CH_{OG}$ & $E_{IGG}$ \\ \hline 
G.PS & \small $R_{GE}^{PS} :C_{A} =C_{B} =2.05$ & \{$\vec{x}_{O}$, $\vec{x}_{R}$, $\vec{x}_{P}$, $\$\vec{x}_{DP}$\} & $\vec{x}_{C} $ & \{$\vec{x}_{O} $,$\vec{x}_{R} $,$\vec{x}_{P}$\} \\ \hline 
G.DE1 & \small $R_{GE}^{DE} :C_{F} =0.5,C_{CR} =0.1,C_{CG} = 1.0$ & \{$\vec{x}_{P} $, $\$\vec{x}_{DP}\}$ & $\vec{x}_{C}$ & $\{\vec{x}_{P}\}$ \\ \hline 
G.DE2 & \small $R_{GE}^{DE} :C_{F} =0.5,C_{CR} =0.9,C_{CG} = 1.0$ & \{$\vec{x}_{P} $, $\$\vec{x}_{DP}\}$ & $\vec{x}_{C}$ & $\{\vec{x}_{P}\}$ \\ \hline 
G.SC & \small $R_{GE}^{SC} :C_{NTB} =2$ & \{$\vec{x}_{R} $, $\$\vec{x}_{GR}$\} & $\vec{x}_{C} $ & $\{\vec{x}_{R} $, $\$\vec{x}_{GR}\}$ \\ \hline 
\end{tabular}
\label{tab:SPEC-G}
\end{table*}

\begin{table*} [!ht]
\centering \caption{Seven CGO cases supported by SPEC-MM, where $E_{UPD} \in E_{IGM} = \bigcup{E_{IGG}} = \{\vec{x}_{O}, \vec{x}_{R}, \vec{x}_{P}, \$\vec{x}_{GR}\}$, and each CGO case has its name $ID_{C}$ (e.g., \#PS) and its set of $C_W$ values in the corresponding column}
\begin{tabular}{|c|c|c|c|c||c|c|c|c|c|c|c|} \hline 
$ID_{G}$ & $\vec{x}_{O} $ & $\vec{x}_{R} $ & $\vec{x}_{P} $ & $\$\vec{x}_{GR}$ & \#PS & \#DE1 & \#DE2 & \#SC & \#DEDE & \#DEPS & \#DESC \\ \hline 
G.PS & $\surd$ & $\surd$ & $\surd$ &  & 1 & 0 & 0 & 0 & 0 & 1 & 0 \\ \hline 
G.DE1 &  &  & $\surd$ &  & 0 & 1 & 0 & 0 & 1 & 0 & 0 \\ \hline 
G.DE2 &  &  & $\surd$ &  & 0 & 0 & 1 & 0 & 1 & 1 & 1 \\ \hline 
G.SC &  & $\surd$ &  & $\surd$ & 0 & 0 & 0 & 1 & 0 & 0 & 1 \\ \hline 
\end{tabular}
\label{tab:SPEC-MM}
\end{table*}

\begin{table*} [!ht]
\centering \caption{The executive rows of SPEC-MM for \#DESC-I, an customized CGO case (``$\surd$'' indicates additional elements)}
\begin{tabular}{|c|c|c|c|c||c|} \hline 
$ID_{G}$ & $\vec{x}_{O} $ & $\vec{x}_{R} $ & $\vec{x}_{P} $ & $\$\vec{x}_{GR}$ & \#DESC-I \\ \hline 
G.PS & $\surd$ & $\surd$ & $\surd$ &  & 0 \\ \hline 
G.DE1 &  &  & $\surd$ &  & 0 \\ \hline 
G.DE2 &  & \underbar{$\surd$} & $\surd$ &  & 1 \\ \hline 
G.SC &  & $\surd$ & \underbar{$\surd$} & $\surd$ & 1 \\ \hline 
\end{tabular}
\label{tab:SPEC-MM-Exp}
\end{table*}

Table \ref{tab:SPEC-MM} lists seven CGO cases. All of them are defined on four executive rows, in which each executive row is associated with a generative row in SPEC-G, according to its $ID_{G}$. Columns 2-5 in Table \ref{tab:SPEC-MM} include the updating lists in $E_{IGM} = \{\vec{x}_{O}, \vec{x}_{R}, \vec{x}_{P}, \$\vec{x}_{GR}\}$. All $E_{SUB}$ lists are defined in the default mode, and the elements in each $E_{SUB}$ (i.e., the corresponding $E_{IGG}$ elements shown in Table \ref{tab:SPEC-G}) are marked by ``$\surd$''. In the last seven columns, the seven CGO cases with different $ID_{C}$, i.e., \#PS, \#DE1, \#DE2, \#SC, \#DEDE, \#DEPS, and \#DESC, are defined by assigning with different $C_{SV}$ values for the four executive rows, in which each executive row is not actually used if the corresponding $C_{SV}$ value is 0. Thus, the first four cases only use a single executive row, whereas the last three cases use two executive rows in the equal probability.

Table \ref{tab:SPEC-MM-Exp} describes the \#DESC-I case, in which only both G.DE2 and G.SC are actually used, and the $E_{SUB}$ lists are specified in the customized mode. Compared to \#DESC, two additional elements, i.e., $\vec{x}_{R} $ for G.DE2 and $\vec{x}_{P} $ for G.SC, are marked by ``\underbar{$\surd$}''. The two executive rows are independent in \#DESC, but cooperative in \#DESC-I. Thus the customized mode provides an additional dimension for further designing new algorithms with additional cooperative rows. Such a cooperation may facilitate for the search process of other embedded search heuristics (ESHs).

Each CGO case is designated by $ID_{C}$ and a few setting parameters. In the current script, there are only two parameters, i.e., the number of agents ($N$) and the maximum number of cycles ($T$).

\subsubsection{Informal Execution Process}

Here an informal description is provided for the actual execution of the \#DESC-I case in Table \ref{tab:SPEC-MM-Exp}. The formal description of the executionc an be found in Section \ref{sec:cgos}. 

For an agent $i$, both rows G.DE2 and G.SC in SPEC-MM (Table \ref{tab:SPEC-MM-Exp}) have the same probability to be selected by $B_{MM(i)}$. Here we assume that G.SC is selected in the current cycle. Thus there are $ID_G$=G.SC, and $E_{UPD}=\{\vec{x}_{R}, \vec{x}_{P}, \$\vec{x}_{GR}\}$. With $ID_G$, $B_{GEN}$ locates the fourth row in SPEC-G (Table \ref{tab:SPEC-G}), and executes the $R_{GE}^{SC}$ instance with $E_{IG}=\{\vec{x}_{R}, \$\vec{x}_{GR}\}$ from $M_{A(i)}$ and $M_S$ and $CH_{OG}=\vec{x}_{C}$ to $M_{G(i)}$. Afterward, the elements in $E_{UPD}$ are processed by $B_{SUB}$, based on the corresponding rows in SPEC-MP (Table \ref{tab:SPEC-MP}). For example, $CH_{U}=\vec{x}_{C}$ is sent to the cell in the buffer $M_{BA(i)}$ that is used for update  $\vec{x}_{P(i)}\in M_{A(i)}$, and $CH_{U}=\vec{x}_{R}$ is sent to the cell in the buffer $M_{BSG}$ that is used for update $\$\vec{x}_{GR} \in M_{SG}$. Note that $M_{BSG}$ might receive chunks from different agents. At the end of the cycle, each buffer cell with new content will be used to update the corresponding LTM cell by the corresponding $R_{UE}$ rule defined in SPEC-MP (Table \ref{tab:SPEC-MP}). 

At next cycles, G.DE2 and G.SC in SPEC-MM might be interactively selected. As we can see, G.SC only use two elements, i.e., $E_{IG}=\{\vec{x}_{R}, \$\vec{x}_{GR}\}$, but it updates three elements, i.e., $E_{UPD}=\{\vec{x}_{R}, \vec{x}_{P}, \$\vec{x}_{GR}\}$. The extra update is marked by ``\underbar{$\surd$}'' in Table \ref{tab:SPEC-MM-Exp}, and it will only have an impact when $\vec{x}_{P}$ serves as an input chunk element for G.DE2.  

\subsection{Discussion}

We have described the solid implementation of various CGO cases. Given the CGO framework, the implementation process including two parts, i.e., realize algorithmic components in the toolbox, and organize the instances of these components by the script. 
There are only a few components in the toolbox. In Table \ref{tab:SPEC-MP}, we have two generic chunk types, i.e.,  $\vec{x}$ and $\$\vec{x}$, a few generic rules, i.e., $R_{IE:X}^{RND}$, $R_{UE:S}^{D}$, $R_{UE:S}^{G}$, and $R_{UE:X}^{TS}$. These generic components might be used across different problem types. In Table \ref{tab:SPEC-G}, we have three problem-specific rules, i.e., $R_{GE}^{PS}$, $R_{GE}^{DE}$, and $R_{GE}^{SC}$.

The memory protocol ontology is defined on a few chunks. Based on the position and corresponding updating process in the updatable graph, nontrivial properties may emerge for these chunks. In the individual memory of each agent, $\vec{x}_{O}$, $\vec{x}_{R}$, and $\vec{x}_{P}$ are the best state found so far, the most recently found state, and the state found in the last cycle, respectively. In the group memory, $\$\vec{x}_{DP}$ is the collection of elite states found by the agents, and $\$\vec{x}_{GR}$ is a steady-state set. 

The primary advantage is to define different apparent stand-alone algorithms in an efficient way. For the ESHs, G.PS is a PSO instance, G.DE1 and G.DE2 are two DE instances, and G.SC is a SCO instance. Different instances of a single rule (e.g., $R_{GE}^{DE}$) might be included in SPEC-E. To develop new algorithms, the main effort is put on realizing $R_{GE}$ with heterogeneous $E_{IG}$ and $CH_{OG}$. Some possible $R_{GE}$ subtypes that include common algorithmic operators (e.g., recombination, mutation, and local search) are discussed in \ref{sec:RGE}. For example, Step $c$ of $R_{GE}^{SC}$ can be viewed as a recombination operator. Furthermore, as discussed in Section \ref{sec:toolbox}, some ESHs might not necessarily be cultural learning strategies (i.e., $R_{GE}$ uses inputs in $M_{A}$ and $M_{S}$). For example, a single-start (e.g., local search and mutation) search rule and the $R_{GE}^{RND}$ rule might be respectively added to Table 
\ref{tab:SPEC-G} using $E_{IG}$=\{$\vec{x}_{P}$\} and $E_{IG}=\varnothing$, given  $CH_{OG}$=$\vec{x}_{C}$. We might also change the realization of an ESH by changing the chunks used in its $E_{IG}$ and $CH_{OG}$. For example, G.SC might turn into a totally different algorithm if it uses the $E_{IG}$ of G.DE2.

Moreover, hybrid CGO cases can be formed in a combinatorial algorithmic space formed by using a portfolio of user-oriented ESHs as basis, without writing any additional code. Here \#DEDE, \#DEPS, and \#DESC, are defined in the default mode, and \#DESC-I is defined in the customized mode. We have only considered the customized mode in an expanded style that adds additional elements into the $E_{UPD}$ lists. The customized mode might be more flexible, as long as it follows the basic principle that any elements in $E_{IGM}$ can be probabilistically updated. The ability of supporting a large algorithmic space offering algorithm designers quick turnaround in realizing hybrid CGO cases. The portfolio design might be guided using the offline performance of individual ESHs accumulated over time.

\section{Experimental Results}
\label{sec:exp}

The experiments are performed on thirteen widely-used benchmark instances ($G$01${\sim}G$13) \cite{Runarsson:2005p1196}
originating from real-world applications. Table \ref{tab:Problems} summarizes the diverse characteristics of the benchmark instances \cite{Runarsson:2005p1196}.
Four of the instances, i.e., $G$03, $G$05, $G$11, and $G$13, have equality constraints. By default, the tolerance value for each equality constraint is $\epsilon_{H}$=1E-4.
memory specification
\begin{table*} [!ht]
\centering \caption{Summary of main characteristics of the benchmark instances \cite{Runarsson:2005p1196}: [linear inequality (LI), nonlinear equality (NE), nonlinear inequality (NI), and the number of active constraints at optimum (NA)]}
\begin{tabular}{|c|c|c|c|c|c|c|c|} \hline 

$F_{P}$ & $D$ & $f(x)$ type & $|S_{PF}|/|S_{P}|$ & LI & NE & NI & NA \\ \hline 
$G$01 & 13 & quadratic & 0.011\% & 9 & 0 & 0 & 6 \\ \hline 
$G$02 & 20 & nonlinear & 99.990\% & 1 & 0 & 1 & 1 \\ \hline 
$G$03 & 10 & polynomial & 0.000\% & 0 & 1 & 0 & 1 \\ \hline 
$G$04 & 5 & quadratic & 52.123\% & 0 & 0 & 6 & 2 \\ \hline 
$G$05 & 4 & cubic & 0.000\% & 2 & 3 & 0 & 3 \\ \hline 
$G$06 & 2 & cubic & 0.006\% & 0 & 0 & 2 & 2 \\ \hline 
$G$07 & 10 & quadratic & 0.000\% & 3 & 0 & 5 & 6 \\ \hline 
$G$08 & 2 & nonlinear & 0.856\% & 0 & 0 & 2 & 0 \\ \hline 
$G$09 & 7 & polynomial & 0.512\% & 0 & 0 & 4 & 2 \\ \hline 
$G$10 & 8 & linear & 0.001\% & 3 & 0 & 3 & 3 \\ \hline 
$G$11 & 2 & quadratic & 0.000\% & 0 & 1 & 0 & 1 \\ \hline 
$G$12 & 3 & quadratic & 4.779\% & 0 & 0 & 9$^3$ & 0 \\ \hline 
$G$13 & 5 & exponential & 0.000\% & 0 & 3 & 0 & 3 \\ \hline
\end{tabular}
\label{tab:Problems}
\end{table*}

The algorithm performance is measured on the mean results under given numbers of function evaluations (NFE).
For each CGO case, its NFE is approximately equal to $N \cdot T$, since the evaluation times in $t$=0 can be neglected if $T$ is large enough. For each problem instance, 500 independent runs were performed for obtaining the mean results. Furthermore, only the runs that entered $S_{PF}$ are taken into accounted, and the number of runs that did not enter $S_{PF}$ is reported in parentheses.

For an algorithm case, a problem instance is regarded as \textit{solved} if the difference between the mean result and the optimal value is smaller than 1E-5 (except for \textit{G}08 and \textit{G}13, which are 1E-6). All the solved results listed in following tables are emphasized in \textbf{boldface}. 
As comparing sub-optimal results with different algorithms (shown in Tables \ref{tab:Comp1}, \ref{tab:Comp2}, and \ref{tab:ExpandedProblems}), each existing result is simply \underline{underlined} if it has no statistically significant difference from the corresponding \#DESC-I result at 95\% confidence level, based on Welch's t-test.

\subsection{Algorithm Selection}
\label{sec:basic}

The selection process is based on the two insights in algorithm portfolio design \cite{Huberman:1997p1159}. 
For all tests in Section \ref{sec:basic}, there are $N$=60 and $T$=2000, and NFE is 1.2E4.

Table \ref{tab:Res_Single} summarizes the mean results by four pure CGO cases, i.e., \#DE1, \#DE2, \#PS, and \#SC, in which each case uses a single ESH. Here some nontrivial knowledge about the competency of the corresponding ESHs may be obtained from their offline performance.

For the instances without equality constraints, the $f$*$^{1}$ values are the true optimal solutions. For those instances with equality constraints, the $f$*$^{1}$ values are the optimal solutions obtained as $\epsilon_{H}$=1E-4.

As shown in Table \ref{tab:Res_Single}, \#DE1, \#DE2, \#PS, and \#SC consistently achieved the optimal solutions in six, eleven, five, and five of the instances, respectively. \#DE2 achieved the best search performance among the four CGO cases. For the two instances $G$01 and $G$02, DE1 and \#SC were able to achieve very good results, whereas \#DE2 and \#PS did not obtain good enough results.

\begin{table*} [!ht]
\centering \caption{Results by four pure CGO cases using a single ESH}
\begin{tabular}{|c|c|c|c|c|c|} \hline 
$F_{P}$ & $f$*$^{1}$ & \#DE1 & \#DE2 & \#PS & \#SC \\ \hline 
$G$01 & -15.00000 & \textbf{-15.00000} & -14.78906 & -14.90595 & \textbf{-15.00000} \\ \hline 
$G$02 & -0.80362 & -0.80091 & -0.62628 & -0.64812 & -0.79764 \\ \hline 
$G$03 & \emph{-1.00050} & -0.99493 & \textbf{-1.00050} & -1.00045 & N/A \\ \hline 
$G$04 & -30665.5387 & \textbf{-30665.5387} & \textbf{-30665.5387} & \textbf{-30665.5387} & \textbf{-30665.5387} \\ \hline 
$G$05 & \emph{5126.49671} & (500) & \textbf{5126.49671} & 5137.3522(8) & N/A \\ \hline 
$G$06 & -6961.81388 & \textbf{-6961.81388} & \textbf{-6961.81388} & \textbf{-6961.81388} & \textbf{-6961.81388} \\ \hline 
$G$07 & 24.30621 & 24.79860 & \textbf{24.30621} & 25.14686 & 24.41236 \\ \hline 
$G$08 & -0.095825 & \textbf{-0.095825} & \textbf{-0.095825} & \textbf{-0.095825} & \textbf{-0.095825} \\ \hline 
$G$09 & 680.63006 & 681.03268 & \textbf{680.63006} & 680.65376 & 680.64244 \\ \hline 
$G$10 & 7049.24802 & 7211.44153 & \textbf{7049.24802} & 7456.49008 & 7166.74353 \\ \hline 
$G$11 & \emph{0.74990} & \textbf{0.74990} & \textbf{0.74990} & \textbf{0.74990} & N/A \\ \hline 
$G$12 & -1.00000 & \textbf{-1.00000} & \textbf{-1.00000} & \textbf{-1.00000} & \textbf{-1.00000} \\ \hline 
$G$13 & \emph{0.053942} & (500) & \textbf{0.053942} & 0.073435 & N/A \\ \hline 
\end{tabular}
\label{tab:Res_Single}
\end{table*}

\begin{table*} [!ht]
\centering \caption{Results by four hybrid CGO cases using two ESHs}
\begin{tabular}{|c|c|c|c|c|c|} \hline 
$F_{P}$ & \#DEDE & \#DEPS & \#DESC & \#DESC-I & $St. Dev.$ \\ \hline 
\emph{}$G$01 & -14.99531 & -14.96719 & \textbf{-15.00000} & -14.99997 & 2.205E-05 \\ \hline 
$G$02 & -0.79712 & -0.69590 & -0.79287 & -0.79006 & 1.255E-02 \\ \hline 
$G$03 & \textbf{-1.00050} & \textbf{-1.00050} & \textbf{-1.00050} & \textbf{-1.00050} & 1.489E-10 \\ \hline 
$G$04 & \textbf{-30665.5387} & \textbf{-30665.5387} & \textbf{-30665.5387} & \textbf{-30665.5387} & 2.942E-10 \\ \hline 
$G$05 & \textbf{5126.49671} & \textbf{5126.49671} & \textbf{5126.49671} & \textbf{5126.49671} & 9.346E-12 \\ \hline 
$G$06 & \textbf{-6961.81388} & \textbf{-6961.81388} & \textbf{-6961.81388} & \textbf{-6961.81388} & 3.277E-11 \\ \hline 
$G$07 & \textbf{24.30621} & \textbf{24.30621} & \textbf{24.30621} & \textbf{24.30621} & 3.305E-07 \\ \hline 
$G$08 & \textbf{-0.095825} & \textbf{-0.095825} & \textbf{-0.095825} & \textbf{-0.095825} & 5.835E-16 \\ \hline 
$G$09 & \textbf{680.63006} & \textbf{680.63006} & \textbf{680.63006} & \textbf{680.63006} & 2.855E-12 \\ \hline 
$G$10 & 7049.24822 & 7049.24812 & 7049.24812 & 7049.24813 & 1.370E-04 \\ \hline 
$G$11 & \textbf{0.74990} & \textbf{0.74990} & \textbf{0.74990} & \textbf{0.74990} & 6.001E-15 \\ \hline 
$G$12 & \textbf{-1.00000} & \textbf{-1.00000} & \textbf{-1.00000} & \textbf{-1.00000} & 0.000E-00 \\ \hline 
$G$13 & \textbf{0.053942} & \textbf{0.053942} & \textbf{0.053942} & \textbf{0.053942} & 2.114E-16 \\ \hline 
\end{tabular}
\label{tab:Res_Multi}
\end{table*}

\begin{table*} [!ht]
\centering \caption{Results for instances with equality constraints as $\epsilon_{H}$=1E-8}
\begin{tabular}{|c|c|c|c|c|c|} \hline 
$F_{P}$ & $f$*$^{2}$ & \#DE2 & \#DESC & \#DESC-I & $St. Dev.$ \\ \hline 
$G$03 & -1.00000 & \textbf{-1.00000} & \textbf{-1.00000} & \textbf{-1.00000} & 1.748E-05 \\ \hline 
$G$05 & 5126.49811 & \textbf{5126.49811} & 5126.50203 & 5126.49812 & 8.143E-05 \\ \hline 
$G$11 & 0.75000 & \textbf{0.75000} & \textbf{0.75000} & \textbf{0.75000} & 4.334E-15 \\ \hline 
$G$13 & 0.053950 & 0.054720 & 0.055489 & \textbf{0.053950} & 2.799E-09 \\ \hline 
\end{tabular}
\label{tab:Res_1E-8}
\end{table*}

Table \ref{tab:Res_Multi} gives the mean results by \#DEDE, \#DEPS, \#DESC, and \#DESC-I, in which each hybrid case employs two ESHs. Here \#DEPS is included since it represents an existing algorithm called DEPSO \cite{Zhang:2003p1404}. For \#DESC-I, the standard deviation ($St. Dev.$) is provided.

All the four hybrid cases achieved the optimal solutions for ten instances, which may mean that they inherited most of the merit from G.DE2. Moreover, they all achieved better results than \#DE2 for both $G$01 and $G$02. The portfolio may benefit from the negative correlation among the performance of individual algorithms \cite{Huberman:1997p1159}. Besides, both DESC and DESC-I performed better results than \#DEDE in $G$01 and $G$10, and \#DEPS in $G$01 and $G$02, respectively. 

For the four instances with equality constraints, Table \ref{tab:Res_1E-8} gives the optimal solutions ($f$*$^{2}$) and the mean results obtained by three CGO cases, i.e., \#DE, \#DESC, and \#DESC-I, as the allowed tolerance value is reduced to $\epsilon_{H}$=1E-8. All the three CGO cases found the optimal solutions for $G$03 and $G$11. Furthermore, \#DE2 and \#DESC-I respectively found the optimal solutions for $G$05 and $G$13. For $G$05, the result of \#DESC-I was slightly worse than the optimal solution, but it achieved a much better result than \#DE2.

For an instance with equality constraints, a smaller $\epsilon_{H}$ leads to a more accurate optimal solution. Compared to the real optimal solutions as $\epsilon_{H}$=0, the $f$*$^{2}$ results are the same, whereas the $f$*$^{1}$ results still have the differences that are not negligible, under the given arithmetic precisions. It is meaningful that \#DESC-I was able to achieve near-optimal solutions for all the four instances, even as the using of a smaller $\epsilon_{H}$ might significantly increase the problem difficulty.

For the instances with equality constraints, $F_{R}$ is adjusted dynamically by using a dependent chunk, i.e., $\$\vec{x}_{DP}$. In \#DESC, only one executive row, i.e., G.DE2, may update $\vec{x}_{P}$, which is a root chunk of $\$\vec{x}_{DP}$, in the $M_{A}$ of each agent. Thus the progress of the other executive row, i.e., G.SC, is totally ignored. Actually, \#DESC did not find better results than \#DE2. However, in \#DESC-I, G.SC updates $\vec{x}_{P}$, while G.DE2 updates $\vec{x}_{R}$ as well. Such a mutual interaction ensures that the search progress of both the executive rows are taken into account. As shown in Table \ref{tab:Res_1E-8}, \#DESC-I was able to achieve a much better result than \#DE2 in $G$13.

The difference between \#DESC and \#DESC-I is in that the two ESHs in \#DESC are independent, whereas in \#DESC-I they are cooperative due to the additional updating elements. Compared \#DESC-I to \#DESC, such an interaction significantly enhanced the performance for $G$05 and $G$13, as shown in Table \ref{tab:Res_1E-8}. Thus, for an algorithm portfolio, the overall performance might be further tuned through low-level cooperative search among individual algorithms \cite{Huberman:1997p1159}. 

\begin{table*} [!ht]
\centering \caption{Mean results by \#DESC-I:S and four existing algorithms}
\begin{tabular}{|c|c|c|c|c|c|c|c|} \hline 
$F_{P}$ & \#DESC-I:S & $St. Dev.$ & GA$_{SAFF}$ & OEA & CDE  & PSO$_{SAV}$ & ES$_{SM}$ \\ \hline
$G$01 & -14.99373 & 3.32E-03 & -14.9993 & \textbf{-15} & \textbf{-14.999996} & -14.715104 & \textbf{-15.000} \\ \hline
$G$02 & -0.76896 & 3.38E-02 & -0.77512 & -0.782518 & -0.724886 & -0.740577 & -0.785238 \\ \hline
$G$03 & \textbf{-1.00050} & 7.64E-04 & -0.99930 & \textbf{-1.000} & -0.788635 & -1.003367 & \textbf{-1.000} \\ \hline
$G$04 & \textbf{-30665.5387} & 8.58E-08 & -30659.41 & \textbf{-30665.539} & \textbf{-30665.539} & \textbf{-30665.538672} & \textbf{-30665.539} \\ \hline
$G$05 & \textbf{5126.49671} & 2.83E-08 & N/A & 5127.048 & 5207.410651 & 5202.362681 & 5174.492 \\ \hline
$G$06 & \textbf{-6961.81388} & 3.28E-11 & -6961.769 & \textbf{-6961.814} & \textbf{-6961.814} & \textbf{-6961.813875} & -6961.284 \\ \hline
$G$07 & 24.30765 & 2.09E-03 & 27.83 & 24.373 & \textbf{24.306210} & 24.988731 & 24.475 \\ \hline
$G$08 & \textbf{-0.095825} & 5.95E-16 & -0.092539 & \textbf{-0.095825} & \textbf{-0.095825} & \textbf{-0.095825} & \textbf{-0.095825} \\ \hline
$G$09 & \textbf{680.63006} & 1.06E-11 & 680.97 & 680.632 & \textbf{680.630057} & 680.655378 & 680.643 \\ \hline
$G$10 & 7049.66674 & 1.27E+00 & 7760.54 & 7219.011 & 7049.248266 & 7173.266104 & 7253.047 \\ \hline
$G$11 & \textbf{0.74990} & 6.00E-15 & 0.7546 & \textbf{0.750} & 0.757995 & \textbf{0.749002} & \textbf{0.75} \\ \hline
$G$12 & \textbf{-1.00000} & 0.00E+00 & -0.99972 & \textbf{-1} & \textbf{-1.000000} & \textbf{-1} & \textbf{-1.000} \\ \hline
$G$13 & 0.055977 & 7.94E-01 & N/A & 0.053969 & 0.288324 & 0.552753 & 0.166385 \\ \hline
\end{tabular}
\label{tab:Comp1}
\end{table*}

\begin{table*} [!ht]
\centering \caption{Mean results by \#DESC-I:L and four existing algorithms}
\begin{tabular}{|c|c|c|c|c|c|c|c|} \hline 
$F_{P}$ & \#DESC-I:L & $St. Dev.$ & SIMP$_{\alpha}$ & ES$_{ATM}$ & ES$_{RY05}$ & SAMO-GA & SAMO-DE \\ \hline
$G$01 & \textbf{-15.00000} & 6.67E-08 & \textbf{-15.00000} & \textbf{-15.000} & \textbf{-15.000} & \textbf{-15.0000} & \textbf{-15.0000} \\ \hline
$G$02 & -0.79080 & 1.10E-02 & -0.78419 & \underline{-0.790148} & -0.782715 & -0.79605 & -0.79874 \\ \hline
$G$03 & \textbf{-1.00000} & 2.84E-07 & \textbf{-1.00050} & \textbf{-1.000} & \textbf{-1.001} & \textbf{-1.0005}  & \textbf{-1.0005} \\ \hline
$G$04 & \textbf{-30665.5387} & 2.94E-10 & \textbf{-30665.5387} & \textbf{-30665.539} & \textbf{-30665.539} & \textbf{-30665.5386} & \textbf{-30665.5386}\\ \hline
$G$05 & \textbf{5126.49811} & 2.29E-11 & \textbf{5126.49671} & 5127.648 & \textbf{5126.497} & 5127.976 & \textbf{5126.497} \\ \hline
$G$06 & \textbf{-6961.81388} & 3.27E-11 & \textbf{-6961.81388} & -\textbf{6961.814} & \textbf{-6961.814} & \textbf{-6961.81388} & \textbf{-6961.81388} \\ \hline
$G$07 & \textbf{24.30621} & 2.24E-10 & 24.30626 & 24.316 & \textbf{24.306} & 24.4113 & 24.3096 \\ \hline
$G$08 & \textbf{-0.095825} & 1.00E-15 & \textbf{0.095825} & \textbf{-0.09825} & \textbf{-0.095825} & \textbf{-0.095825} & \textbf{-0.095825} \\ \hline
$G$09 & \textbf{680.63006} & 2.91E-12 & \textbf{683.63006} & 683.639 & \textbf{680.630} & 683.634 & \textbf{680.630} \\ \hline
$G$10 & \textbf{7049.24802} & 3.28E-08 & \textbf{7049.24802} & 7250.437 & 7049.250 & 7144.40311 & 7059.81345 \\ \hline
$G$11 & \textbf{0.75000} & 4.22E-15 & \textbf{0.74990} & \textbf{0.75} & \textbf{0.750} & \textbf{0.7499} & \textbf{0.7499} \\ \hline
$G$12 & \textbf{-1.00000} & 0.00E+00 & \textbf{1.00000} & \textbf{-1} & \textbf{-1.000} & \textbf{1.0000} & \textbf{1.0000} \\ \hline
$G$13 & \textbf{0.053950} & 4.78E-16 & 0.066770 & 0.053959 & 0.066770 & 0.054028 & \textbf{0.053942} \\ \hline
\end{tabular}
\label{tab:Comp2}
\end{table*}

\subsection{Comparison with Existing Algorithms}

The performance of two CGO cases, i.e., \#DESC-I:S and \#DESC-I:L, was compared to that of existing algorithms. For the two CGO cases, \#DESC-I:S has $N$=50 and $T$=1000, and thus its NFE is 5.0E4; while \#DESC-I:L has $N$=70 and $T$=3000, and thus its NFE is 2.1E5. 

\subsubsection{Existing Algorithms}
\label{sec:ExsitAlgs}

Here we briefly describe basic features of ten algorithms, including their diverse algorithmic types and constraint-handling techniques (some of them can be represented by $R_{QC}$ rules discussed in Section \ref{sec:QCRules}). For more details, please refer to corresponding literature.

GA$_{SAFF}$ \cite{Farmani:2003p1127} is a GA with the self-adaptive fitness formulation. Specifically, each individual is assigned an infeasibility value, i.e., the normalized sum of all constraint violation values, and the two-stage penalty functions are then applied in relation to boundary solutions.

OEA \cite{Liu:2007p1475} is an organizational evolutionary algorithm. The individuals in a population are structured into some organizations, in which all evolutionary operations are applied for simulating the interaction among the organizations. For handling constraints, the static penalty term $R_{QC}^{P}$, which uses the penalty coefficient tuned for each problem instance, is considered.

CDE is a cultured differential evolution \cite{Becerra:2006p1496}, i.e., a CA algorithm hybridized with a DE population, or a DE algorithm integrated with a belief space of CA. In CDE, CA uses the external knowledge in its belief space to influence the DE search operations. Its constraint-handling method can be regarded as the natural quality measurement using the $R_{QC}^{O}$ rule.

PSO$_{SAV}$, or called SAVPSO \cite{Lu:2008p1507}, is a PSO variant, in which each particle adjusts its velocity self-adaptively, according to the run-time information, for searching within $S_{PF}$. Its constraint-handling method can be described by the quality measurement using $R_{QC}^{O}$.

SIMP$_{\alpha}$, or called $\alpha$Simplex \cite{Takahama:2005p1387}, is an $\alpha$-constrained simplex method. The $\alpha$-constrained method uses a satisfaction level in order to indicate how well a search point (state) satisfies the constraints. The simplex method uses multiple simplexes to avoid the situation that a single simplex might lose its affine independence. For each simplex, the worst point is mutated, either by a boundary mutation if the point is feasible, or by another mutation to increase the satisfaction level of the point. The $\alpha$-level is dynamically increased for the problem instances with equality constraints. 

ES$_{ATM}$, or called ATMES \cite{Wang:2008p1549}, integrates a ($\mu$, $\lambda$)-ES with a constraint-handling method called adaptive tradeoff model, which tries to achieve a good tradeoff between feasible and infeasible spaces during different stages of a search process, by taking advantage of the valuable run-time information. For equality constraints, the dynamic adjustment used in \cite{Hamida:2002p1205} is adopted.

ES$_{SM}$ is a simple multimembered ES \cite{MezuraMontes:2005p1325}, a ($\mu$+$\lambda$)-ES variant with two basic modifications, i.e., a panmictic combined recombination technique to improve its exploitation capability and a diversity mechanism that keeps best infeasible solutions in the population. Its constraint-handling method can be seen as using $R_{QC}^{O}$ for the quality measurement. For equality constraints, the dynamic adjustment used in \cite{Hamida:2002p1205} is considered. 

ES$_{RY05}$ \cite{Runarsson:2005p1196} is a ($\mu$, $\lambda$)-ES variant hybridized with differential variation. Its constraint-handling method is the stochastic ranking technique, which is equivalent to applying the $R_{QC}^{SR}$ rule for quality measurement. 

SAMO-GA and SAMO-DE \cite{Elsayed2011} are two self-adaptive multi-operator (SAMO) based algorithms that using multiple genetic and DE operators, respectively. Each search operator has its own sub-population, and the sub-populations are changed by a self-adaptive learning strategy during the evolution process.  

\subsubsection{Comparison}

Table \ref{tab:Comp1} and \ref{tab:Comp2} list the results obtained by \#DESC-I:S and \#DESC-I:L, and the existing algorithms. 

For the instances with equality constraints, the tolerance value $\epsilon_{H}$=1E-4 was considered in all the experiments, except for the experiments by \#DESC-I:L, ES$_{ATM}$, ES$_{SM}$, and PSO$_{SAV}$, which used $\epsilon_{H}$= 1E-8, 5E-6, 4E-4, and 1E-3, respectively.

The NFEs of ES$_{ATM}$, ES$_{SM}$, and OEA were 2.4E5. The NFEs of ES$_{RY05}$ and GA$_{SAFF}$ were 3.5E5. The NFEs of SIMP$_{\alpha}$, CDE, and PSO$_{SAV}$ were 2.9E5$\sim$3.3E5, 1.0E5, and 5.0E4, respectively.

For NFE, \#DESC-I:L is less than ES$_{ATM}$, ES$_{SM}$, ES$_{RY05}$, GA$_{SAFF}$, OEA, and SIMP$_{\alpha}$, while \#DESC-I:S is further less than or similar to CDE and PSO$_{SAV}$. 

As for consistently achieving optimal solutions, the two cases \#DESC-I:S and \#DESC-I:L were successful in eight and twelve instances, while ES$_{ATM}$, ES$_{SM}$, ES$_{RY05}$, GA$_{SAFF}$, OEA, SIMP$_{\alpha}$, CDE, and PSO$_{SAV}$ were successful in seven, six, ten, zero, seven, ten, seven, and five instances, respectively. For $G$02, \#DESC-I:L performed better than all the existing algorithms.

Compared to PSO$_{SAV}$, \#DESC-I:S was dominating. Furthermore, \#DESC-I:S outperformed ES$_{SM}$, OEA, GA$_{SAFF}$, and CDE, although it spent a much less NFE. Compared to EA$_{SM}$, \#DESC-I:S achieved better results in six instances ($G$05, $G$06, $G$07, $G$09, $G$10, and $G$13), and similar results in five instances. \#DESC-I:S dominated GA$_{SAFF}$ in all the instances, except $G$02. Compared to OEA, \#DESC-I:S achieved better results in four instances ($G$05, $G$07, $G$09, and $G$10), and similar results in six instances. Compared to CDE, \#DESC-I:S achieved better results in five instances ($G$02, $G$03, $G$05, $G$11, and $G$13), and similar results in five instances.

Other five algorithms are compared to \#DESC-I:L. For ES$_{ATM}$, the tolerance value for the instances with equality constraints is $\epsilon_{H}$=5E-6. As demonstrated in Table \ref{tab:Comp2}, \#DESC-I:L worked efficiently on harder instances with $\epsilon_{H}$=1E-8. Compared to ES$_{ATM}$ on the other nine instances, \#DESC-I:L achieved better results in $G$02, $G$07, $G$09, and $G$10, and obtained similar results in five easy instances. Compared to ES$_{RY05}$, \#DESC-I:L achieved better results in $G$02, $G$10, and $G$13, and found similar results in all other instances. Compared to SIMP$_{\alpha}$, \#DESC-I:L performed better in $G$02, $G$07, and $G$13, and found similar results in all other instances. Compared to SAMO-GA, \#DESC-I:L performed better in five instances and worse in $G$02. Compared to SAMO-DE, \#DESC-I:L performed better in $G$07 and $G$10 and worse in $G$02. 

Overall, \#DESC-I:L found better results over the existing algorithms on the classic benchmark set, with a less NFE (except for CDE and PSO$_{SAV}$) and a tighter $\epsilon_{H}$ for problems with equality constraints.

\subsection{Additional Tests}

In this section, additional tests are performed for demonstrating the run-time behavior and the role of landscape tuning of the hybrid CGO case \#DESC-I:L.  

\subsubsection{Run Length Distribution}

\begin{figure} [!ht]
\centering \includegraphics[width=3.3in]{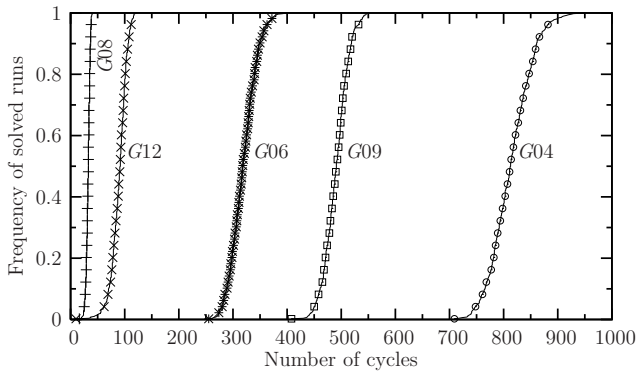} \caption{RLDs for \#DESC-I:L on \textit{G}04, \textit{G}06, \textit{G}08, \textit{G}09, \textit{G}12.}
\label{Fig:RLD_s}
\end{figure}

\begin{figure} [!ht]
\centering \includegraphics[width=3.3in]{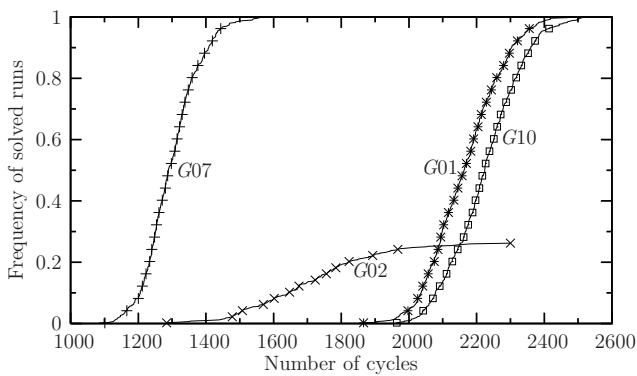} \caption{RLDs for \#DESC-I:L on \textit{G}01, \textit{G}02, \textit{G}07, and \textit{G}10.}
\label{Fig:RLD_l}
\end{figure}

\begin{figure} [!ht]
\centering \includegraphics[width=3.3in]{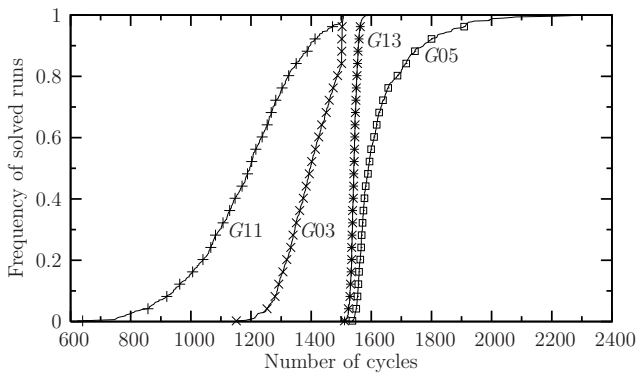} \caption{RLDs for \#DESC-I:L on \textit{G}03, \textit{G}05, \textit{G}11, and \textit{G}13.}
\label{Fig:RLD_e}
\end{figure}

The run-length distribution (RLD) \cite{Hoos2004} is suitable for characterizing the run-time behavior of a stochastic algorithm case on each problem instance. Figures \ref{Fig:RLD_s} to \ref{Fig:RLD_e} show the frequency of solved runs along with the number of cycles for \#DESC-I:L on all thirteen instances. All the instances were 100\% solved, except for \textit{G}02. The steepness of each RLD discloses nontrivial information for the variance of the run length of each algorithm case. For the problem instances with equality constraints, their RLDs are influenced by the threshold cycle, i.e., $c_{TH}$=$C_{RTU} \cdot T$=1500. Here, \textit{G}03 and \textit{G}11 were mostly solved before $c_{TH}$, whereas \textit{G}05 and \textit{G}13 were solved after $c_{TH}$ and most of the solutions were obtained quite soon after $c_{TH}$, although there is a longer tail for \textit{G}05. For all instances without equality constraints, their RLDs are steep, except for \textit{G}02. Since the success frequency is sufficiently high, using independent runs can further improve the performance \cite{Hoos2004}. 

RLD gives a quite accurate estimation of competency of an stochastic algorithm. In addition, RLD might be easily approximated using a Weibull distribution \cite{Hoos2004} with only two parameters. In other words, the information of algorithm behavior might be compressed and stored for possible analysis and learning.

\subsubsection{Landscape Tuning}

In the facilitator, $R_{ADJ}^{RR}$ essentially encoding the knowledge on handling equality constraints for adaptively shaping the problem landscape.  
Because $\$\vec{x}_{FB}=\$\vec{x}_{DP}$ is a run-time chunk in $M_{S}$, only three other parameters might be used for tuning the landscape. Note that $R_{ADJ}^{RR}$ does not account for landscape tunings on the problem instances without equality constraints.

Tables \ref{tab:Res_CRRE} to \ref{tab:Res_CRTU} report the average results for \#DESC-I:L using different $C_{RRE}$, $C_{RNU}$, and $C_{RTU}$ values. The algorithm achieved optimal results in a large parameter space. $C_{RRE}$ should not be too large, since the relaxation loses its usage. $C_{RNU}$ should not be close to 1 to ensure that Eq. \ref{eq:ERE} is executed. $C_{RTU}$ should not be too small or too large. Before $t_{TH} ={\rm INT}(C_{RTU} \cdot T )$, the adaptive ratio-reaching technique plays the role for ensuring a suitable pace to reach the neighborhood of the optimal solution; after $t_{TH}$, $R_{QC}^{RRR}$ is equivalent to $R_{QC}^{O}$, which is useful for fine search.

We expect that any global features that can explicitly tuning the problem landscape might be encoded in the group facilitator. For a better understanding the capability of each $R_{M}$ rule, it is certainly useful to analyze the influence for different parameter combinations on different problems. It might be more robust and effective to use an ensemble of constraint handling techniques \cite{Mallipeddi2010b}, including those $R_{QC}$ rules in Section \ref{sec:QCRules}. 

Furthermore, the essential role of landscape tuning is only providing transformed problem instances. The majority of efforts for achieving better performance should still be placed on the customization in the algorithmic space formed from the portfolio of ESHs. 

\begin{table*} [!ht]
\centering \caption{Results for \#DESC-I:L using different $C_{RRE}$ values on the instances with equality constraints}
\begin{tabular}{|c|c|c|c|c|c|} \hline 
$F_{P}$ & $C_{RRE}$=1E0 & $C_{RRE}$=1E1 & $C_{RRE}$=1E2 & $C_{RRE}$=1E3 & $C_{RRE}$=1E4 \\ \hline 
$G$03 & \textbf{-1.00000} & \textbf{-1.00000} & \textbf{-1.00000} & -0.99991 & -0.99860 \\ \hline 
$G$05 & \textbf{5126.49811} & \textbf{5126.49811} & \textbf{5126.49811} & 5126.49823 & 5126.49917 \\ \hline 
$G$11 & \textbf{0.75000} & \textbf{0.75000} & \textbf{0.75000} & \textbf{0.75000} & \textbf{0.75000} \\ \hline 
$G$13 & \textbf{0.053950} & \textbf{0.053950} & \textbf{0.053950} & \textbf{0.053950} & 0.054721 \\ \hline 
\end{tabular}
\label{tab:Res_CRRE}
\end{table*}

\begin{table*} [!ht]
\centering \caption{Results for \#DESC-I:L using different $C_{RNU}$ values on the instances with equality constraints}
\begin{tabular}{|c|c|c|c|c|c|} \hline 
$F_{P}$ & $C_{RNU}$=0.00 & $C_{RNU}$=0.25 & $C_{RNU}$=0.50 & $C_{RNU}$=0.75 & $C_{RNU}$=1.00 \\ \hline 
$G$03 & \textbf{-1.00000} & \textbf{-1.00000} & \textbf{-1.00000} & \textbf{-1.00000} & -0.03707(414) \\ \hline 
$G$05 & \textbf{5126.49811} & \textbf{5126.49811} & \textbf{5126.49811} & 5126.50580 & 5313.93466 \\ \hline 
$G$11 & \textbf{0.75000} & \textbf{0.75000} & \textbf{0.75000} & \textbf{0.75000} & 1.00707(40) \\ \hline 
$G$13 & \textbf{0.053950} & \textbf{0.053950} & \textbf{0.053950} & \textbf{0.053950} & (500) \\ \hline 
\end{tabular}
\label{tab:Res_CRNU}
\end{table*}

\begin{table*} [!ht]
\centering \caption{Results for \#DESC-I:L using different $C_{RTU}$ values on the instances with equality constraints}
\begin{tabular}{|c|c|c|c|c|c|} \hline 
$F_{P}$ &$C_{RTU}$=0.00 &$C_{RTU}$=0.25 &$C_{RTU}$=0.50 &$C_{RTU}$=0.75 &$C_{RTU}$=1.00 \\ \hline 
$G$03 & -0.96073 & -0.99699 & \textbf{-1.00000} & \textbf{-1.00000} & \textbf{-1.00000} \\ \hline 
$G$05 & 5126.49823 & 5126.49812 & \textbf{5126.49811} & \textbf{5126.49811} & (500) \\ \hline 
$G$11 & \textbf{0.75000} & \textbf{0.75000} & \textbf{0.75000} & \textbf{0.75000} & \textbf{0.75000}  \\ \hline 
$G$13 & 0.085691 & 0.055809 & \textbf{0.053950} & \textbf{0.053950} & (500) \\ \hline 
\end{tabular}
\label{tab:Res_CRTU}
\end{table*}

\subsubsection{Expanded Problem Set}

We continue to evaluate on an expanded problem set from the CEC06 competition \cite{Liang2006}. For end users, this might be viewed as the situation of their problems changing over time. Table \ref{tab:ExpandedProblems} give the results by \#DESC-I:L, SAMO-GA, and SAMO-DE. Here \#DESC-I:L is running at $N=80$. Thus for all the algorithms, NFE=2.4E5. 

\begin{table*} [!ht]
\centering \caption{Results by \#DESC-I:L, \#DESC-I:L2, SAMO-GA, and SAMO-DE}
\begin{tabular}{|c|c|c|c||c|c|c||c|c|} \hline 
$F_{P}$ & \#DESC-I:L & $St. Dev.$ & $Median$ & \#DESC-I:L2 & $St. Dev.$ & $Median$ & SAMO-GA & SAMO-DE   \\ \hline 
$G$14 & \textbf{-47.76489} & 5.00E-15 & \textbf{-47.76489} & \textbf{-47.76489} & 3.00E-15 & \textbf{-47.76489} & -46.47318 & -47.68115  \\ \hline 
$G$15 & \textbf{961.71502} & 5.00E-15 & \textbf{961.71502} & \textbf{961.71502} & 0.00E+00 & \textbf{961.71502} & 961.71509 & \textbf{961.71502}   \\ \hline 
$G$16 & \textbf{-1.905155} & 0.00E+00 & \textbf{-1.905155} & \textbf{-1.905155} & 0.00E+00 & \textbf{-1.905155} & -1.905154 & \textbf{-1.905155}   \\ \hline 
$G$17 & 8887.26933 & 4.17E+01 & \textbf{8853.53967} & \underline{8891.23276} & 4.24E+01 & \textbf{8853.53967} & 8853.8871 & \textbf{8853.5397}   \\ \hline 
$G$18 & -0.82591 & 7.79E-02 & \textbf{-0.866025} & -0.85036 & 5.25E-02 & \textbf{-0.866025} & -0.865545 & -0.866024  \\ \hline 
$G$19 & 32.6912 & 8.52E-02 & 32.66595 & \textbf{32.65559} & 2.01E-07 & \textbf{32.65559} & 36.427463 & 32.75734  \\ \hline 
$G$21 & 202.98569 & 2.96E+01 & \textbf{193.72451} & 193.76478 & 8.90E-01 & \textbf{193.72451} & 246.09154 & 193.77138  \\ \hline 
$G$23 & -382.93683 & 4.86E+01 & -400.02517 & \underline{-385.49695} & 6.42E+01 & \textbf{-400.05510} & -194.76034 & -360.81766  \\ \hline 
$G$24 & \textbf{-5.508013} & 0.00E+00 & \textbf{-5.508013} & \textbf{-5.508013} & 0.00E+00 & \textbf{-5.508013} & \textbf{-5.508013} & \textbf{-5.508013}   \\ \hline 
\end{tabular}
\label{tab:ExpandedProblems}
\end{table*}

Compared to SAMO-GA, \#DESC-I:L performed better in six instances and only worse in $G$17 and $G$18. Compared to SAMO-DE,  \#DESC-I:L achieved better results in three instances ($G$14, $G$19, $G$23), and comparable results in three instances ($G$15, $G$16, and $G$24). In addition, the median solutions of \#DESC-I:L achieved or approached closely to the optimal solutions.

Previous CGO cases only use naive search operators, for the sake of simplicity in description. The performance might be benefited from other superior operators. In Table \ref{tab:ExpandedProblems}, \#DESC-I:L2 is slightly tuned from \#DESC-I:L by using $C_{CG}$=0.5 for G.DE2. In other words, the DE operator is tuned from best/2/bin into current-to-best/2/bin, where ``current-to-best'' is a more advanced sub-strategy in DE operators \cite{Zhang2009,Elsayed2013}. Compared to \#DESC-I:L, this minor change achieved statistically significant improvement on three instances ($G$18, $G$19, and $G$20).  In addition, all median solutions of \#DESC-I:L2 achieved the optimal solutions.

The advance of \#DESC-I:L2 might be largely attributed to better preservation of the population diversity, since ``current-to-best'' relies less on the currently best state. The cooperative group itself provides an implicit mechanism to preserve diversity in the memory of agents. However, it is still beneficial to prevent states becoming too similar. In a human idea-generating group, redundant states are often automatically filtered. Thus, CGO cases might be further improved by considering some explicit strategies to control redundant states, e.g., the speciation method and $R_{GE}^{RND}$ used in multimethod \cite{Vrugt2009} and multi-operator algorithms \cite{Elsayed2011,Elsayed2012,Elsayed2013}.

In addition, the CGO system might also be benefited from including advanced search operators (e.g., those used in SAMO-DE) to further improve the performance on some problem instances, e.g., $G$17 and $G$18.

\subsection{Summary}

The algorithm selection process was consistent with two nontrivial insights in algorithm portfolio design \cite{Huberman:1997p1159,Streeter:2008p1072}. First, combining competent strategies into a portfolio may improve the overall performance by exploiting the negative correlation among the offline performance of individual strategies (e.g., \#DE2 and \#SC). Second, customized cooperative search (\#DESC-I) further tuned the portfolio performance than when the individual strategies are independent (\#DESC). The two insights might help users to quickly find an effective algorithm case in a large algorithmic space.

We then compared the performance of \#DESC-I with existing algorithms in diverse paradigms and with different constraint-handling techniques. With less computational times, CGO cases achieved competitive performance as compared to existing algorithms. 

The behavior of \#DESC-I was demonstrated using the run length distributions. We also performed a systematic test to show the impacts in tuning the problem landscape. An expanded problem set were also tested, and a slightly tuned CGO case was also demonstrated.

\section{Related Work and Discussion}
\label{sec:RelWork}

In Section \ref{sec:Characteristics}, we have discussed related work on two main elements, i.e., the cooperative group paradigm and the algorithm portfolio design, of the CGO system. The implementation itself, shows the relation with existing algorithm paradigms, e.g., PSO, DE, and SCO, and the role of shaping landscape in handling constraints. The essential role of the CGO system, however, is a development framework for realizing customized algorithm portfolios with low-level hybridization.

Many conceptual frameworks \cite{Raidl2006,Talbi:2002p1393,Milano:2004p1345,Taillard2001} have been proposed for formalizing (hybrid) metaheuristics. In adaptive memory programming (AMP) \cite{Taillard2001} and a refined version called the multiagent metaheuristic architecture (MAGMA) \cite{Milano:2004p1345}, metaheuristics are defined as a set of algorithmic components operating on a shared memory. They provide a unified view for various metaheuristics, e.g., iterative LS, ACO, and GA. Using a single memory, however, limits the capability to support a cooperative group. There is also a unified view \cite{Raidl2006} that hybrid metaheuristics can be built on a toolbox of components, based on a common pool template. In \cite{Talbi:2002p1393}, a taxonomy has been proposed to distinguish low and high levels, and relay and teamwork models for hybrid metaheuristics. The hybridization among stand-alone algorithms are often limited in the high level, and heterogeneous algorithms can only independently run or light-weighted cooperate through a shared medium. From this taxonomy, hybrid CGO cases belong to a mix of low and high levels, since individual ESHs can work in a self-containing way, while the cooperative search at the low level is emphasized. The CGO framework not only supports simple relay and teamwork models based on the individual and/or group memory, but also the low-level cooperation among heterogeneous ESHs on customized memory elements. In addition, the memory protocol for individual and group memory provides a natural support for quarantining detrimental results.

Various software frameworks have been proposed for realizing hybrid metaheuristics. As surveyed in \cite{Parejo2012}, 33 software frameworks have been identified, and 10 of them have been selected for comparison. Typical examples include HeuristicLab, ParadisEO, and JCLEC, etc. Here we discuss the relations on toolbox elements, and design of pure and hybrid metaheuristics.

These frameworks contain a toolbox of low-level components for the code reuse. In ParadisEO, low-level components are defined as $Helpers$ in some categories, i.e., evolutionary helpers (e.g., transformation, selection, and replacement operations), local search helpers (e.g., generic and problem-specific classes to local search metaheuristics), and some special helpers (e.g., the management of parallel and distributed models). In HeuristicLab, the toolbox contains atomic $operators$ (e.g., mutation, crossover, and selection operators) working on data structures called $scopes$, and micro operators might be defined using a operator graph. In JCLEC, individuals and elementary operations are represented by $IIndividual$ and $ITool$ interfaces (e.g., $IProvider$, $ISelector$, $IRecombinator$ and $IMutation$ respectively for individual initialization, selection, recombination and mutation operations). 
The toolbox in the CGO system follows the same principle for the code reuse. There are chunks and four primary rule interfaces ($R_M$, $R_{IE}$, $R_{UE}$, and $R_{GE}$). Association interfaces are provided in macro components. For example, $R_{SEL}$ might be used in $R_{UE:X}$ and $R_{GE}$, and $R_{QC}$ and $R_{ADJ}$ are used in some macro $R_{M}$ rules. The $R_{GE}$ rule provides a unified entrance for various kinds of search operators, e.g., the operators extracted from DE, PSO, and SCO. Typical operators, e.g., local search and recombination, might be provided in macro $R_{GE}$ rules. Heterogeneous  $R_{GE}$ rules might also be implemented using arbitrary input/output chunks. Generic low-level hybridization models, e.g., relay \cite{Talbi:2002p1393}, can be implemented in macro rules. All these low-level features might be realized in existing software frameworks, but a unified $R_{GE}$ interface provides a basic step to support heterogeneous ESHs. In addition, the use of $R_{M}$ provides a way to utilize the global knowledge on the problem landscape, and might relieve the requirement and difficulty of realizing too many competent components in the toolbox.

Algorithm paradigms are then realized using low-level components. In JCLEC, Algorithms are coded from the $ISystem$ and $IAlgorithm$ interfaces. In ParadisEO, The run of the metaheuristics are coded into $Runners$ by invoking the helpers to perform specific actions on their data, and pure and hybrid metaheuristics are coded into $Solvers$. In HeuristicLab, some algorithms might be designed by using an operator graph, and complex algorithms are created by writing code. Each of these algorithm paradigms might use a configuration file to define parameters and some operators of specific interfaces. These frameworks support different paradigms, e.g., stochastic local search, GA, PSO, and DE, but these algorithms are coded separately. Each algorithm paradigm can only support a very limited algorithmic space to limit the flexibility to adapt to user-specific problem sets, and the configuration file can only provide very limited operator replacements. Hybridization between paradigms often needs to extensively write code by realizing a new paradigm. Thus, there is often a steep learning curve for existing frameworks \cite{Parejo2012}. 

For the CGO system, the framework can be seen as a concrete and unified algorithm paradigm, and the script is its configuration file. The setting in a cooperative group with flexible memory protocol facilitates the realization of heterogeneous ESHs. The effort of designing a stand-alone algorithm (e.g., PSO, DE, and SCO) is reduced to implement a few algorithmic components and then organize them into an ESH using the script. Furthermore, hybridization can be realized by only defining customized algorithm portfolios in the CGO script, without writing any additional code. Based on the insights from algorithm portfolio design, the capability to adapt to specific problem sets might be benefited from negative correlations among the offline performance of individual ESHs. If user-specific problem sets change over time, new heuristics that mainly tackle some of new problems unsolved by existing heuristics can be added into the portfolio, and obsolete ESHs can be removed, if necessarily. There are two kinds of users for a CGO system. Advanced users might focus on realizing algorithmic components in the toolbox (and even improving the CGO framework for highly advanced users), and then designing new ESHs using the script. Basic users might only need pay attention to design customized portfolios based on offline performance of ESHs, and do not need advanced knowledge on algorithmic components and framework details. 

It is also interesting to discuss about the similarities and differences with existing multimethod \cite{Vrugt2009}, multi-operator  \cite{Elsayed2011,Elsayed2012,Elsayed2013}, and ensemble methods \cite{Mallipeddi2010,Mallipeddi2010a,Mallipeddi2010b}. These algorithms provides high-performance realizations in combining the strengths of different search operators (e.g., ES, GA, DE, and PSO) and constraint-handling techniques. These operators might be significant sources for the CGO toolbox. Furthermore, a distinguished feature of these algorithms is that multiple operators compete the offspring generation in sub-populations with adaptively varied sizes, to favor some operators that exhibit higher reproducive success, over the evolution process. Compared to these methods, the CGO system provide a flexible way to support customized low-level cooperation among heterogeneous ESHs that interact on multiple chunks in the private memory of individual agents. 

\subsection{Possible Extensions}
\label{sec:CGO_Ext}

To reduce the complexity for description, the current CGO system is realized in the naive form. 
For future work, some possible extensions are discussed.

Many extensions do not require any changes in the framework. As shown in previous tests, the system can benefit from advanced algorithmic components that tailor to specific problem structures. For large-scale problems, many practical ESHs might only work on partial problems. The agents might still benefit from their cooperative search if the partial problems handled by different ESHs are sufficiently overlapped (otherwise agents might only rely on local reactive strategies \cite{Liu2002}). 

It is also useful to formally define commonly-used macro components for simplifying the reusing of algorithmic components and implementing new ESHs. 

Some extensions require more or less changes on the CGO framework. Here we list two possible extensions, then discuss possible changes on the framework. 

First, the facilitator might provide multiple $R_M$ rules to support an ensemble of constraint handling techniques\cite{Mallipeddi2010b}. Second, the manual process used in the experiments might be regarded as a greedy maximum set cover strategy, and the online selection of ESHs in the portfolio is purely proportionally, where individual ESHs might be favored by providing relatively larger $C_W$ values. It is also possible to utilize some intelligent algorithm selection methods \cite{Xie:2004p1408,Birattari:2002p1068,SmithMiles:2008p1392}), by inferring from the integration of run-time information and offline performance of individual ESHs in the portfolio. 

To support online self-adaptation, the basic intuition \cite{Vrugt2009,Mallipeddi2010b,Elsayed2011} is to favor operators that contribute more for the overall performance. The CGO framework might incorporate three basic changes: 1) each agent associates newly generated chunks with some tags. The tags might include the identifiers of components in usage (e.g., ESH and $R_M$). 2) the facilitator provides some evaluations \cite{Vrugt2009,Mallipeddi2010b,Elsayed2011} on the contributions of the chunks associated with different tags. 3) Based on the evaluation information, each agent updates the weights for algorithmic components in usage. These changes, nevertheless, might only cause few changes in the script.

\section{Conclusions}
\label{sec:conclusion}

The cooperative group optimization (CGO) system is a generic framework to combine the advantages of the cooperative group and low-level portfolio design for realizing CGO algorithms. In the cooperative group, the agents not only search the problem landscape in a parallel way by using diverse knowledge in their individual memory, but also cooperate with their peers by diffusing novel knowledge through the group memory. The search process might also be accelerated through adaptive landscape shaping by a passive group leader. 

The CGO framework is driven by the CGO script that assembles algorithmic components in the CGO toolbox. The CGO script is based on a multilayer design to facilitate the accumulation of knowledge over time. Based on the CGO system, implementing a stand-alone embedded search heuristic (ESH) only needs to implement a few algorithm components, and customized portfolios can be defined using the script without writing any additional code. 
Possible cooperative portfolios form a large algorithmic space. The offline performance of individual ESHs might be used for help adapting the framework to user-specific problem sets, based on the nontrivial insights from algorithm portfolio design. 
Advanced users might focus on realizing algorithmic components and then defining new ESHs using the script, whereas basic users might only pay attention to design customized portfolios on existing ESHs. 

Finally, we have discussed several aspects of the proposed CGO system that warrant further investigation.

\bibliographystyle{spmpsci}

\end{document}